\documentclass[10pt,journal,compsoc]{IEEEtran}
\usepackage{xcolor}
\usepackage{graphicx}
\usepackage{tikz-qtree,ulem}
\usepackage{url}
\usepackage{adjustbox}
\usepackage{subcaption}

\begin{document}

\title{Synthetic Data in Human Analysis: A Survey}
%
%
% author names and IEEE memberships
% note positions of commas and nonbreaking spaces ( ~ ) LaTeX will not break
% a structure at a ~ so this keeps an author's name from being broken across
% two lines.
% use \thanks{} to gain access to the first footnote area
% a separate \thanks must be used for each paragraph as LaTeX2e's \thanks
% was not built to handle multiple paragraphs
%
%
%\IEEEcompsocitemizethanks is a special \thanks that produces the bulleted
% lists the Computer Society journals use for "first footnote" author
% affiliations. Use \IEEEcompsocthanksitem which works much like \item
% for each affiliation group. When not in compsoc mode,
% \IEEEcompsocitemizethanks becomes like \thanks and
% \IEEEcompsocthanksitem becomes a line break with idention. This
% facilitates dual compilation, although admittedly the differences in the
% desired content of \author between the different types of papers makes a
% one-size-fits-all approach a daunting prospect. For instance, compsoc 
% journal papers have the author affiliations above the "Manuscript
% received ..."  text while in non-compsoc journals this is reversed. Sigh.

\author{Indu Joshi, Marcel Grimmer, Christian Rathgeb, Christoph Busch, Francois Bremond,
        Antitza~Dantcheva
\IEEEcompsocitemizethanks{\IEEEcompsocthanksitem I. Joshi, F. Bremond, and A. Dantcheva are with the STARS team of Inria, Sophia Antipolis - Méditerranée and Université Côte d'Azur, France.\protect\\
% note need leading \protect in front of \\ to get a newline within \thanks as
% \\ is fragile and will error, could use \hfil\break instead.
E-mail: indu.joshi@inria.fr, francois.bremond@inria.fr, antitza.dantcheva@inria.fr
\IEEEcompsocthanksitem Marcel Grimmer is with the NBL - Norwegian Biometrics Laboratory, Norwegian University of Science and Technology, Norway.\protect\\
% note need leading \protect in front of \\ to get a newline within \thanks as
% \\ is fragile and will error, could use \hfil\break instead.
E-mail: marceg@ntnu.no
\IEEEcompsocthanksitem C. Rathgeb and C. Busch are with the da/sec - Biometrics and Internet-Security Research Group, Hochschule Darstadt, Germany.\protect\\
% note need leading \protect in front of \\ to get a newline within \thanks as
% \\ is fragile and will error, could use \hfil\break instead.
E-mail: christian.rathgeb@h-da.de, christoph.busch@h-da.de

}
%\IEEEcompsocthanksitem J. Doe and J. Doe are with Anonymous University.}% <-this % stops an unwanted space

%\thanks{Manuscript received ; revised }
}

\IEEEtitleabstractindextext{%
\begin{abstract}
Deep neural networks have become prevalent in human analysis, boosting the performance of applications, such as biometric recognition, action recognition, as well as person re-identification. However, the performance of such networks scales with the available training data. In human analysis, the demand for large-scale datasets poses a severe challenge, as data collection is tedious, time-expensive, costly and must comply with data protection laws. Current research investigates the generation of \textit{synthetic data} as an efficient and privacy-ensuring alternative to collecting real data in the field. This survey introduces the basic definitions and methodologies, essential when generating and employing synthetic data for human analysis. We conduct a survey that summarises current state-of-the-art methods and the main benefits of using synthetic data. We also provide an overview of publicly available synthetic datasets and generation models. Finally, we discuss limitations, as well as open research
problems in this field. This survey is intended for researchers and practitioners in the field of human analysis.
\end{abstract}

% Note that keywords are not normally used for peerreview papers.
\begin{IEEEkeywords}
Human Analysis, Deep Neural Networks, Synthetic Data, Survey
\end{IEEEkeywords}}

% make the title area
\maketitle

% To allow for easy dual compilation without having to reenter the
% abstract/keywords data, the \IEEEtitleabstractindextext text will
% not be used in maketitle, but will appear (i.e., to be "transported")
% here as \IEEEdisplaynontitleabstractindextext when the compsoc 
% or transmag modes are not selected <OR> if conference mode is selected 
% - because all conference papers position the abstract like regular
% papers do.
\IEEEdisplaynontitleabstractindextext
% \IEEEdisplaynontitleabstractindextext has no effect when using
% compsoc or transmag under a non-conference mode.

% For peer review papers, you can put extra information on the cover
% page as needed:
% \ifCLASSOPTIONpeerreview
% \begin{center} \bfseries EDICS Category: 3-BBND \end{center}
% \fi
%
% For peerreview papers, this IEEEtran command inserts a page break and
% creates the second title. It will be ignored for other modes.
\IEEEpeerreviewmaketitle

\IEEEraisesectionheading{\section{Introduction}\label{sec:introduction}}

\IEEEPARstart{W}{e} have witnessed remarkable advancement of deep neural networks (DNNs) in the past decade, leading to mature and robust algorithms in visual perception, natural language processing, and robotic control \cite{lecun2015deep}, among others. Such advancement has been fuelled by the development of \textit{algorithms} to train DNNs, the availability of \textit{large-scale} training \textit{datasets}, as well as the progress in \textit{computational power}.

%Among numerous other domains, 
DNN techniques have been designed, among other applications, for \textit{human analysis}, aiming to recognize human characteristics, behaviour, and interactions with the physical world. In this context, human analysis ranges from the unique authentication of single individuals, the classification of human attributes or actions to the evaluation of crowd-based data. Despite the immense benefit of processing human data, \textit{lack of annotated training data} still hinders DNNs from unfolding their full potential. In addition, the implementation of data protection laws, such as the \textit{European general data protection regulation (GDPR)}, defines strict rules for processing data that can reveal identity information, thus violating the data subjects' informational self-determination. According to article 9 of the GDPR, biometric data is considered as \textit{sensitive data}, and processing without explicit consent of the data subjects is imposed with fines of up to 20 million Euro or 4\% of the firm's worldwide annual revenue from the preceding financial year (article 83).

\begin{figure}
\centering
  \begin{subfigure}[b]{0.29\linewidth}
    \includegraphics[width=\linewidth]{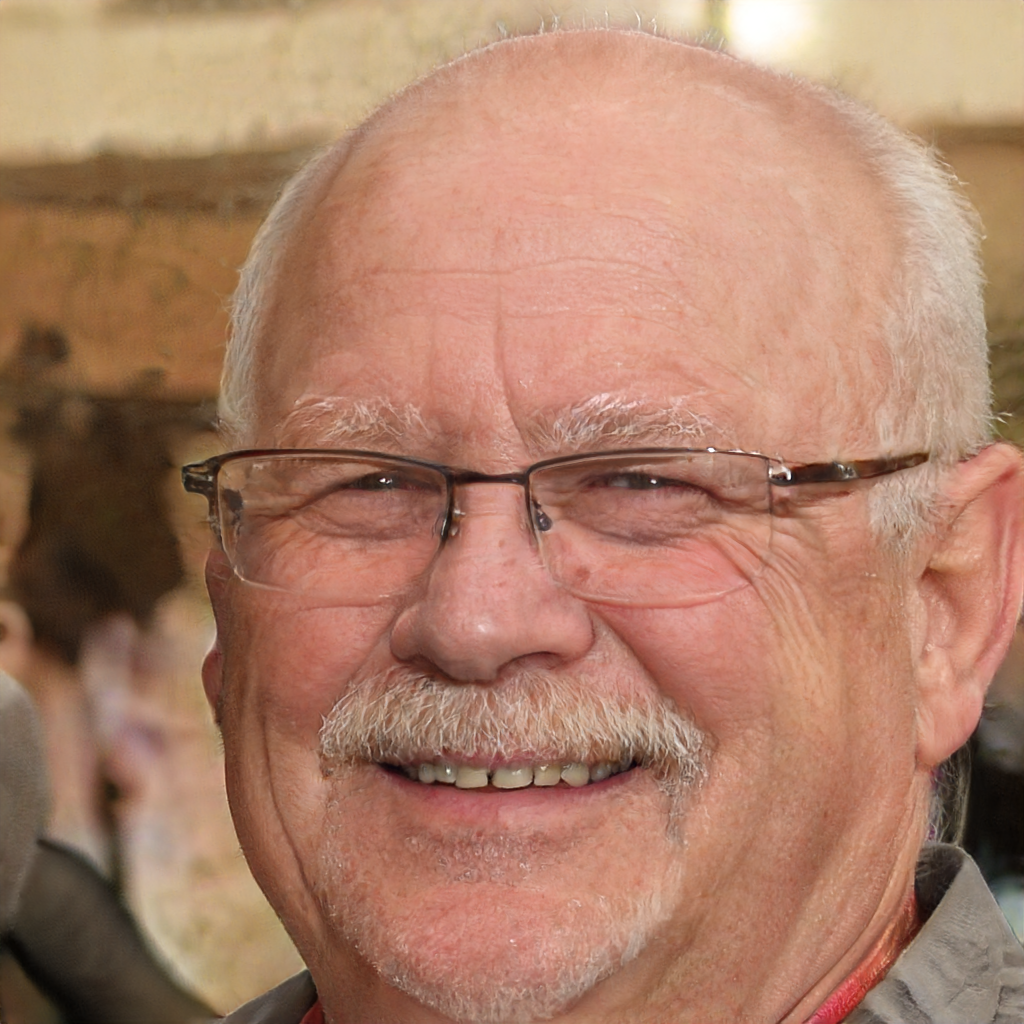}
    \caption{StyleGAN2 \cite{karras2020analyzing}}
    \label{fig:synth-face-example}
  \end{subfigure}
    \begin{subfigure}[b]{0.29\linewidth}
    \includegraphics[width=\linewidth]{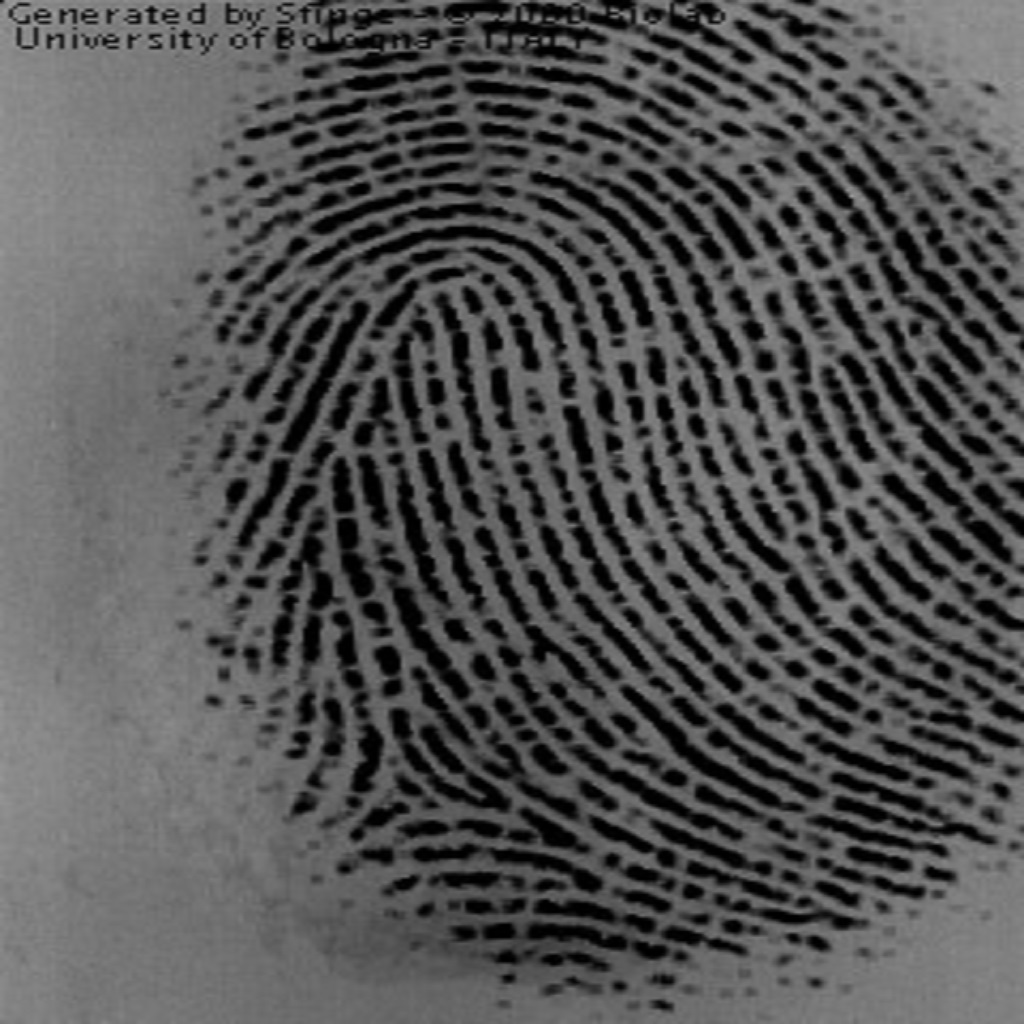}
    \caption{SFinGe \cite{maltoni2009synthetic}}
    \label{fig:synth-face-example}
  \end{subfigure}
  \begin{subfigure}[b]{0.29\linewidth}
    \includegraphics[width=\linewidth]{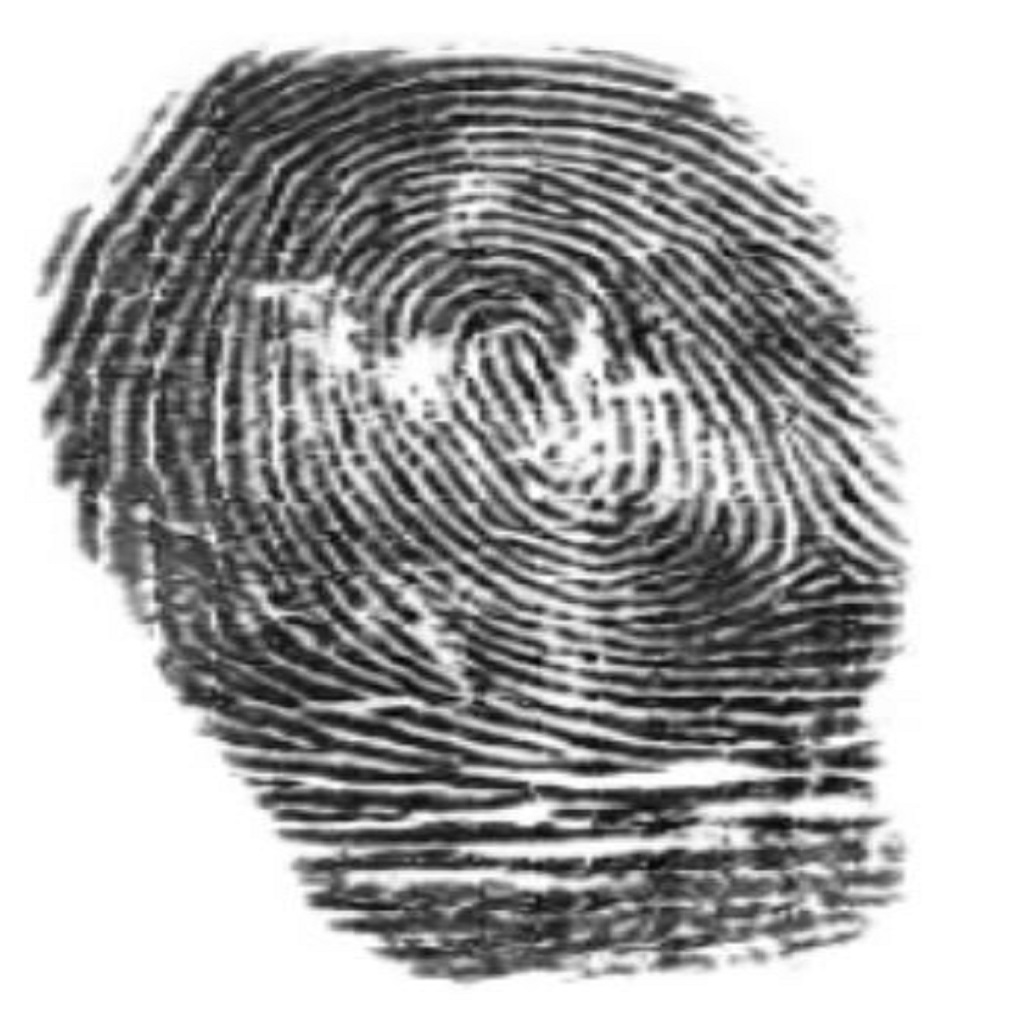}
    \caption{SpoofGAN \cite{grosz2022spoofgan}}
    \label{fig:real-face-example}
  \end{subfigure}\vspace{0.3cm}
  
    \begin{subfigure}[b]{0.29\linewidth}
    \includegraphics[width=\linewidth]{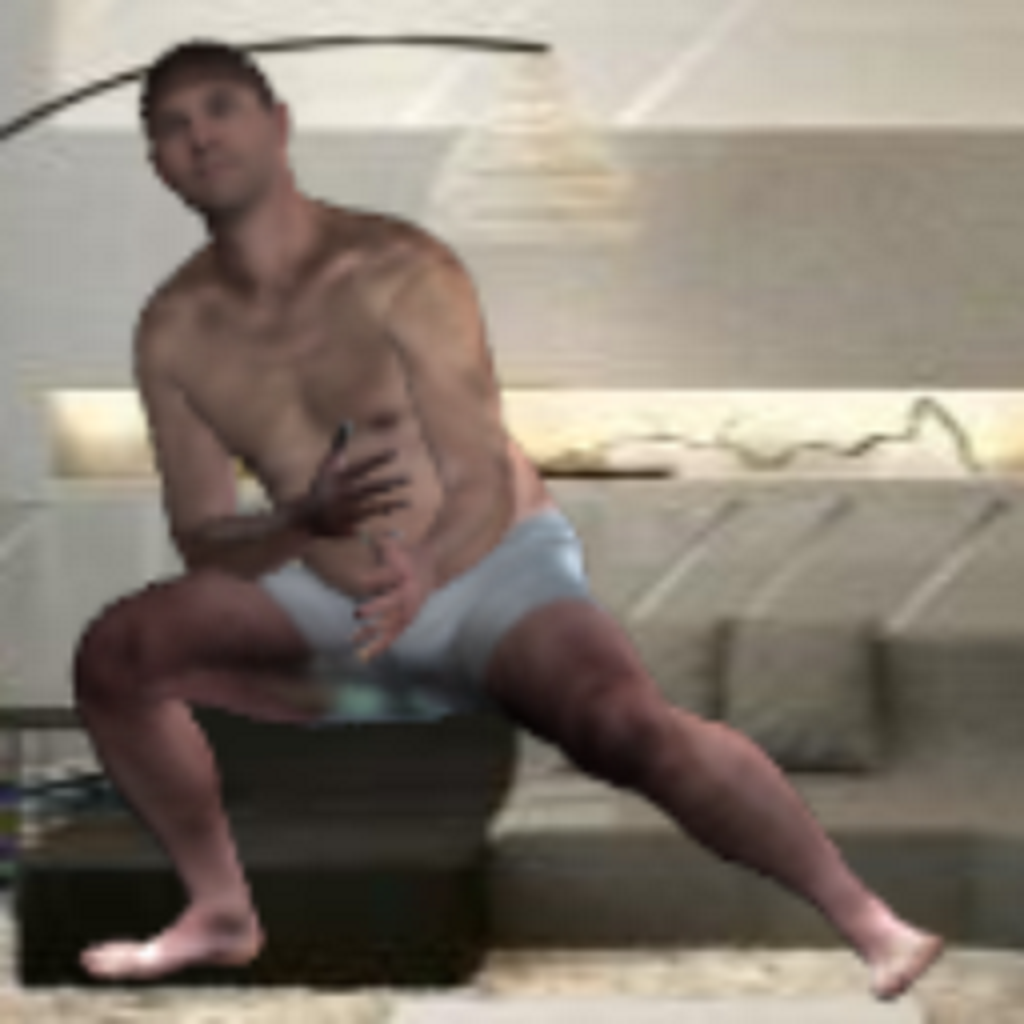}
    \caption{SURREAL \cite{varol2017learning}}
    \label{fig:real-face-example}
  \end{subfigure}
    \begin{subfigure}[b]{0.29\linewidth}
    \includegraphics[width=\linewidth]{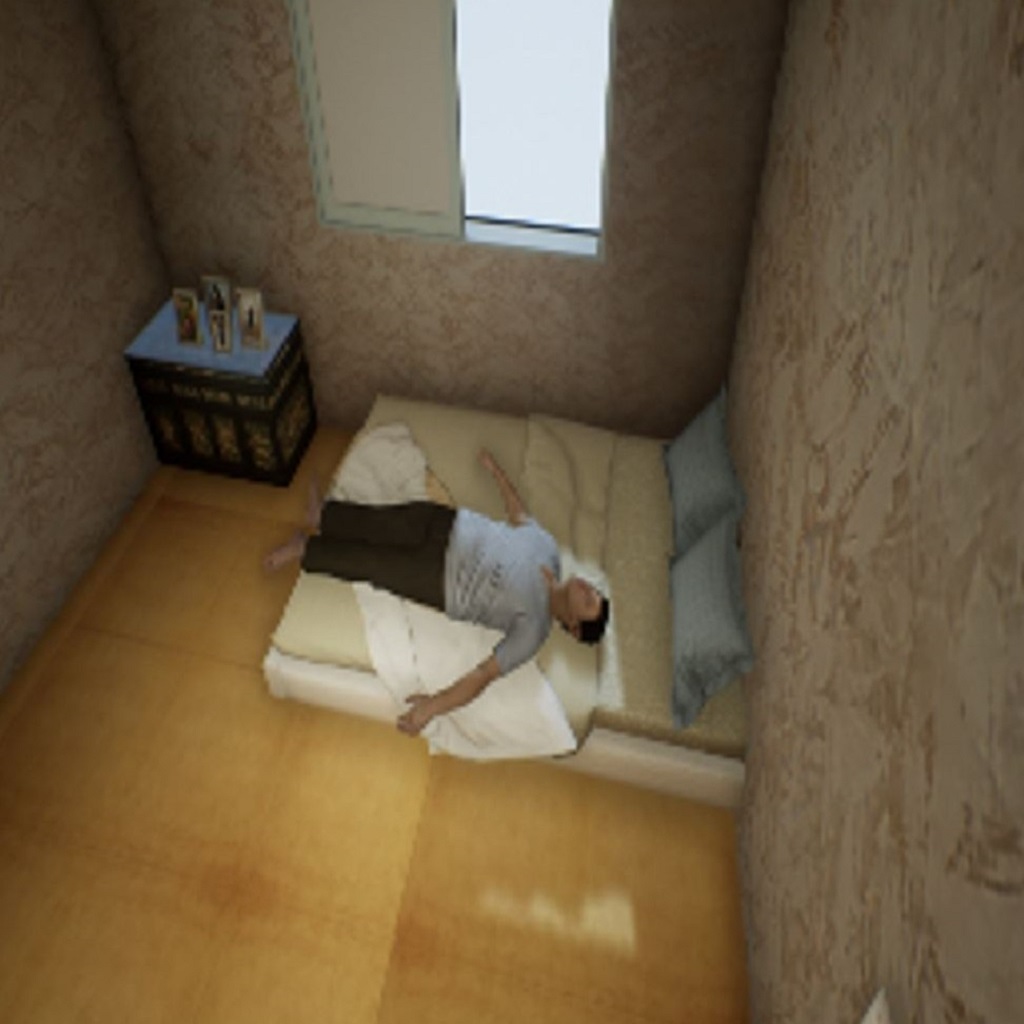}
    \caption{ElderSim \cite{hwang2020eldersim}}
    \label{fig:real-face-example}
  \end{subfigure}
  \caption{Synthetic images generated for human analysis, namely (a) 2D face image generation, (b) fingerprint image generation, (c) fingerprint presentation attack detection, (d) 2D pose estimation, (e) and elderly action recognition   \label{fig:synth-intro-examples}}
\end{figure}

One solution to overcome such limitations related to limited training data and data protection has to do with creating large-scale \textit{synthetic datasets}. Progress of deep generative models has allowed for the generation of highly realistic synthetic human images - challenging to distinguish from real data by both humans, and computer vision algorithms \cite{karras2021alias}\cite{gal2021swagan} (see Figure \ref{fig:synth-intro-examples}). 
While generative models have been able to produce highly realistic synthetic samples, we note that they are prone to leak information from training datasets. This is specifically of concern when human data is involved, and hence identity leaks have to be taken into account to protect personal privacy rights. In this context, current research indicates that identity leaks in deep generative networks become less likely, in case the complexity of the training dataset exceeds the complexity of the model architecture\cite{feng2021gans}. The main reason for identity leaks stems from generative model overfitting to the training dataset, with the consequence of specific units in the network revealing information of single data subjects - a concept referred to as \textit{generative adversarial network (GAN) memorization}.

% Synth vs Real samples

\subsection{Domains of application}
Synthetic data boosts the performance of many data-driven models in human analysis \cite{dou2021versatilegait} \cite{wood2021fake} \cite{irtem2019impact}. In this context, a number of training schemes have been introduced including \textit{data replacement} and \textit{data enrichment}. The motivation for replacing real samples with synthetic data (\textit{i.e.}, \textit{synthetic training}) has to do with alleviating privacy concerns. In contrast, the combination of synthetic and real data (\textit{i.e.}, \textit{augmented training}) mainly aims at reducing biases achieved by re-balancing according to observed soft characteristics. Another optimization scheme aims at initializing model weights based on synthetic data with subsequent fine-tuning on a small subset of real data, referred to as \textit{model initialization}. Finally, domain translation techniques are utilized to close the synthetic vs real domain gap (\textit{domain adaptation}), thereby increasing the realism of synthetic datasets while preserving fine-grained annotations.

Deviating from synthetic data employed for model training, \textit{synthetic evaluation datasets} have been utilized to benchmark the performance of existing algorithms, pre-trained models, and systems. This field of research is fuelled by the increasing representativeness of synthetically generated samples, which allows interference with systems and observed outcomes similar to those expected by real evaluation datasets. The preparation of large-scale testing databases intends to detect weaknesses in the human analysis pipeline without requiring expensive data collection initiatives. Apart from the cost factor, real data from specific (demographic) subgroups may not be accessible, so synthetic samples could fill this gap on a large scale.     

\subsection{Structure of paper}
Given the increasing popularity of synthetic data, the main contribution of this survey is to revisit current research in human analysis, illustrating applications, benefits, and open challenges to accelerate future research. We introduce basic \textit{terminology} and \textit{scope} in Section \ref{sec:fundamentals}, followed by Section \ref{sec:benefits}, which provides an overview of the main \textit{benefits} associated to synthetic data. Section \ref{sec:generation} elaborates on \textit{techniques for generating synthetic data}, followed by the most prominent \textit{application scenarios} presented in Section \ref{sec:usage}. Section \ref{sec:synthDatasets} summarises \textit{synthetic datasets} and \textit{data generation tools} that are publicly available across human analysis domains. Finally, in Section \ref{sec:challenges} we discuss \textit{open challenges} identified in the literature analysis with promising new DNN concepts outlined in Section \ref{sec:conclusion}.

%\IEEEPARstart{W}{e} have witnessed in the past decade remarkable advancement of deep learning (DL), leading to mature and robust algorithms in visual perception, natural language processing, and robotic control \cite{lecun2015deep}, bringing to the fore a large interest and investment. Such advancement has been fuelled by the triptych (a) development of algorithms to train DL-models (e.g., backpropagation), (b) availability of large training datasets (e.g., ImageNet, with 1.4 million labeled images), as well as (c) progress in computational power (e.g., powerful graphical processing units (GPUs)).

%Despite the advancement of DL, we have that current deep neural networks (DNNs) impart mainly limitations related to the necessity for large amounts of annotated training data \cite{de2021next,baradad2021learning}. \textit{Synthetic data} is a solution for this limitation. It is able to circumvent privacy concerns (e.g., legislation in Europe, GDPR), is able to provide a large amount of training samples to optimize millions of network-parameters in data-hungry DNNs, it can facilitate scalability, can provide supervision for training, as well as minimize time and cost of tedious data collection and annotation.

%Survey:  \cite{wang2020survey} 

\section{Synthetic data in human analysis}
\label{sec:fundamentals}

The vast progress of deep generative networks has brought to the fore highly realistic synthetic data beneficial in automated human-centred analysis. To avoid ambiguity throughout this survey and prepare the reader for the following content, we establish terminology of basic concepts and terminology used in this survey.

\subsection{Synthetic data}

In general, \textit{synthetic data} can be defined as \textit{digital information generated by computer algorithms to approximate information collected or measured in the real world} \cite{dankar2021fake}. Synthetic data stems generally from \textit{traditional modelling} or \textit{deep generative models}. While traditional modelling generates real-world patterns based on prior expert knowledge through the \textit{formulation of mathematical models}, deep generative models are designed to \textit{automatically} learn patterns from the training dataset. In the last decade, deep generative models have outperformed traditional modelling techniques, \textit{w.r.t.} quality and generalizability of the synthetic samples \cite{karras2017progressive} \cite{riazi2020synfi}\cite{yadav2019synthesizing}. In this survey, we refer to \textit{generative models} in the context of both mathematical modelling and deep generative models.

\begin{figure}
\centering
  \begin{subfigure}[b]{0.3\columnwidth}
    \includegraphics[width=\linewidth]{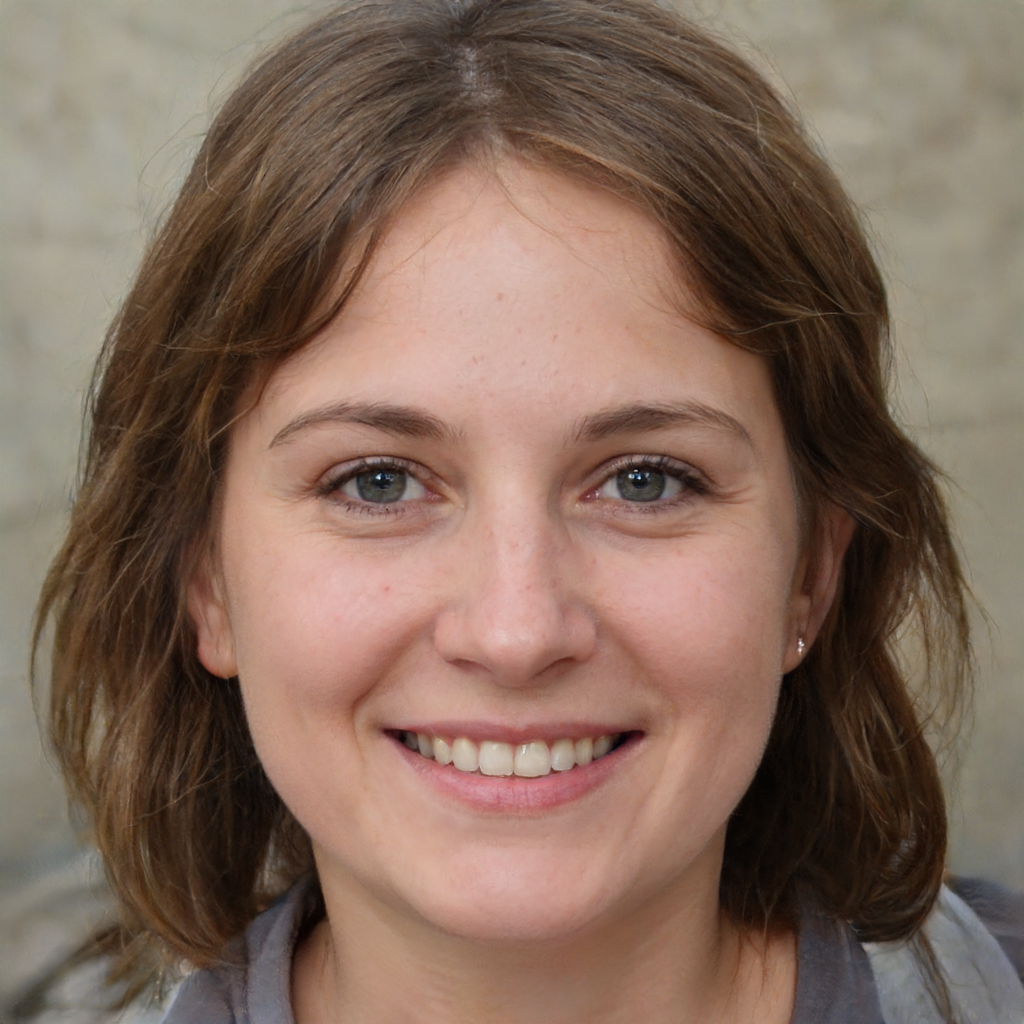}
    \caption{F-S}
    \label{fig:synth-face-example}
  \end{subfigure}
    \begin{subfigure}[b]{0.3\columnwidth}
    \includegraphics[width=\linewidth]{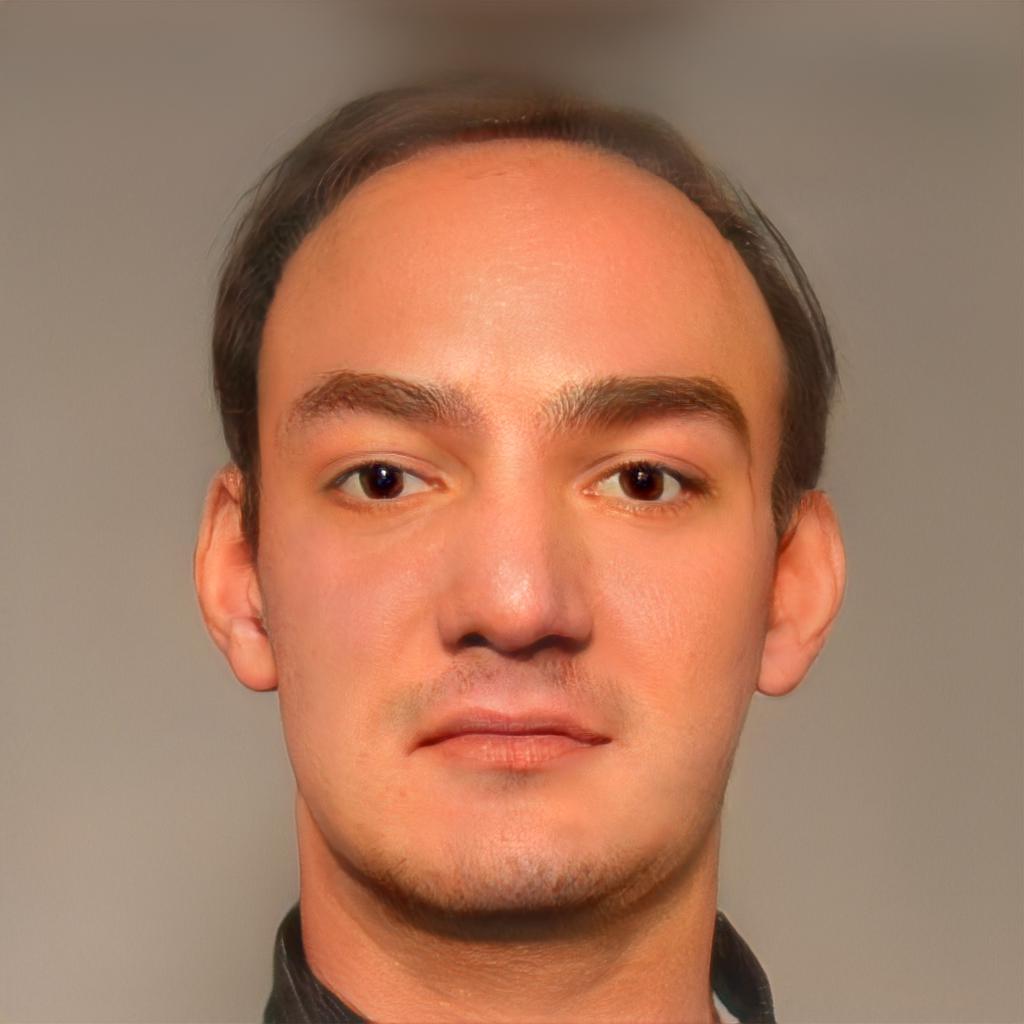}
    \caption{S-S}
    \label{fig:synth-face-example}
  \end{subfigure}
  \begin{subfigure}[b]{0.3\columnwidth}
    \includegraphics[width=\linewidth]{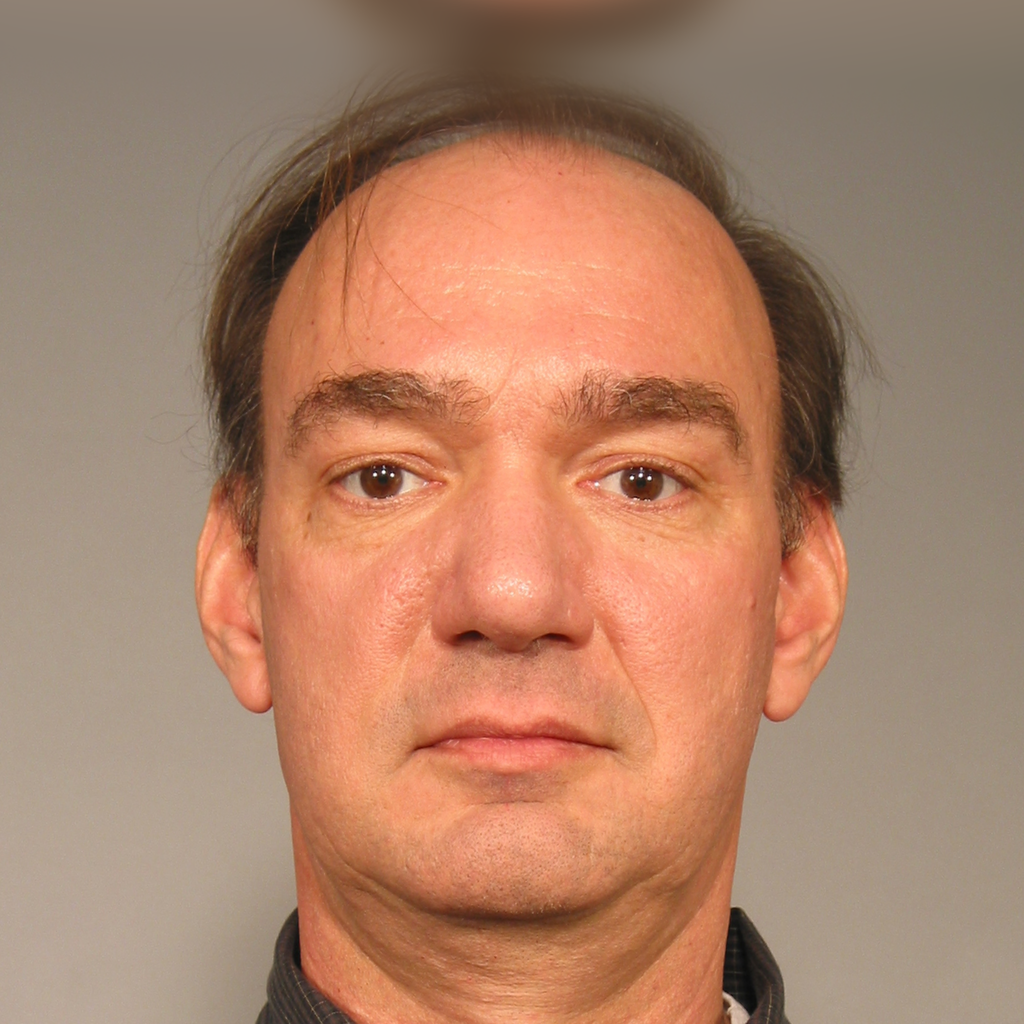}
    \caption{Real}
    \label{fig:real-face-example}
  \end{subfigure}
  \caption{Example images of a fully-synthetic (F-S), semi-synthetic (S-S), as well as real samples. The S-S face image (b) was generated with InterFaceGAN \cite{shen2020interpreting} by editing the age of the real face image depicted on the right side \cite{phillips2005overview}. The F-S sample (a) was randomly generated with StyleGAN2 \cite{karras2020analyzing}.\label{fig:synth-vs-real-example}}
\end{figure}

Synthetic data samples can be \textit{fully-synthetic}, as well as  \textit{semi-synthetic}. Fully-synthetic samples are generated without representing an underlying real-world object \cite{nikolenko2021synthetic}, generally by generative models, random sampling from a learned distribution \cite{engelsma2022printsgan}\cite{karras2019style}. At the same time, semi-synthetic samples constitute representations of real subjects, whose semantics have been manipulated \cite{alaluf2021only}\cite{joshi2021fdeblur}. For example, in human analysis, predicting the future appearance of a real face is considered semi-synthetic, as the image maintains the identity information while altering the age. In contrast, fingerprint images synthesized by GANs based on random noise vectors are defined as fully-synthetic. An example image for each class is demonstrated in Figure \ref{fig:synth-vs-real-example}.

% Synth vs Real samples

In computer vision, real-world information is represented either at \textit{sample} or \textit{feature} level. In particular, we refer to data samples as the analogue or digital representation of human characteristics before feature extraction. According to the harmonic biometric vocabulary of ISO/IEC 2382-37:2017 \cite{ISO-IEC-2382-37-170206}, a feature vector is composed of \textit{numbers or labels extracted from the data sample}. Specifically, feature vectors are treated as compressed sample representations, often encapsulating information optimised for a specific downstream task, such as biometric recognition. In practice, generative models can either focus on generating ``\textit{synthetic samples}'' \cite{karras2021alias}  or ``\textit{synthetic features}''\cite{drozdowski2017sic}, depending on the target application.

\subsection{Data replacement versus Data enrichment}

While deep neural networks have achieved remarkable results in various computer vision tasks, it is still challenging to unleash their full potential due to the limited availability of large-scale datasets. Generation of synthetic samples can improve scalability and diversity, motivated by the following. Firstly, existing datasets being enriched with synthetic samples can increase dataset diversity. % of different factors of variation. 
In this context, \textit{data enrichment (DE)} imparts balancing of the proportions of soft characteristics in order to reduce dataset biases \cite{kortylewski2019analyzing}. Note that in this survey, data enrichment signifies minor \textit{data perturbations} such as image cropping, colour transformation, as well as noise injection \cite{shorten2019survey}. Due to plethora of data augmentation techniques, distinction between \textit{synthetic} and \textit{augmented} samples is often challenging. Therefore, we refer to augmented samples as semi-synthetic, given that the original sample is at hand. In addition, we here denote weak supervision learning as a type of DE, as both synthetic and real samples are jointly employed for model training (see \ref{sec:weak-supervised-learning}).

Secondly, \textit{data replacement (DR)} refers to the replacement of real data with synthetic data \cite{qiu2021synface}. This is instigated by \textit{privacy} concerns in human analysis, where identity information can be linked with the corresponding sample. 

Training human analysis models on domain-adapted synthetic datasets is considered a sub-category of DR, as only high-level information from a small subset of real data is being utilised (see Section \ref{sec:unsupervised-domain-adaption}). In contrast, the initialisation or fine-tuning of model weights with synthetic data is defined as a sub-category of DE due to the active involvement of real data that remains part of the training process (see Section \ref{sec:pretrain-deep-model} and Section \ref{sec:fine-tune-pre-trained-model}). 

%The main principles behind the above-described usage types are summarised as follows.
We proceed to enlist mechanisms in which synthetic data has been employed in the context of DNNs:
\begin{itemize}
    \item \textbf{Augmented Training} refers to learning human analysis models or classifiers from a mixed training dataset that includes both real and synthetic data samples. 
    \item \textbf{Weakly-Supervised Learning} signifies combined training with weak labels (real data) and accurate annotations (synthetic data).
    \item \textbf{Model Initialisation} denotes initial training on synthetic data with subsequent fine-tuning on real data towards reduction of the \textit{synthetic versus real} domain gap. 
    \item \textbf{Model Fine-Tuning} refers to initial training on real data with subsequent fine-tuning on synthetic data to increase the model robustness against biases.
    \item \textbf{Consistency Regularisation} denotes the utilisation of semi-synthetic data to enforce the consistency of model predictions for similar training samples. 
    \item \textbf{Synthetic Training} signifies the training of models or classifiers on datasets composed of synthetic data only.
    \item \textbf{Domain Adaptation} denotes the employment of models trained on synthetic data to domain adaptation techniques (\textit{e.g.,} Cycle-GAN), aiming to close the \textit{synthetic versus real} domain gap.
    \item \textbf{Synthetic Performance Evaluation} refers to assessing synthetic datasets generated to test the scalability and performance of systems, algorithms, or pre-trained models.
\end{itemize}

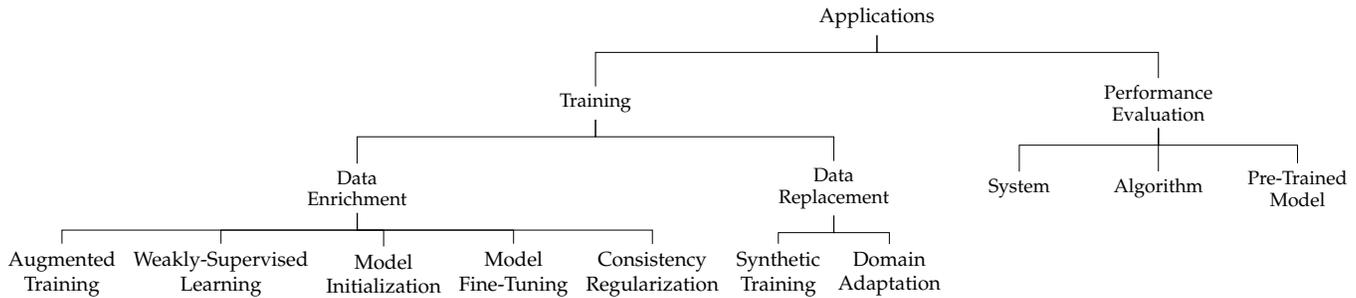
\begin {figure*}%[!hbtp]
\resizebox{\linewidth}{!}{%
\centering
\begin{tikzpicture}[level distance=40pt]
  \tikzset{edge from parent/.style={draw,-, 
    edge from parent path={(\tikzparentnode.south) -- +(0,-8pt) -| (\tikzchildnode)}}}
  \tikzset{every tree node/.style={align=center}}
  \tikzset{every level 1 node/.style={font=\small, text width=2cm}}
  \tikzset{every level 2 node/.style={font=\small, text width=2cm}}

  \Tree [.{Applications}
    [.{Training} [.{Data \\ Enrichment} [.{Augmented \\ Training} ]
                                                      [.{Weakly-Supervised \\ Learning} ][.{Model \\ Initialization} ][.{Model \\ Fine-Tuning} ][.{Consistency \\ Regularization} ]]
                            [.{Data \\ Replacement} [.{Synthetic \\ Training} ]
                                                      [.{Domain \\ Adaptation} ] ]]
    [.{Performance \\ Evaluation} [.{System} ]
                            [.{Algorithm} ][.{Pre-Trained Model} ] ]]
    ]
\end{tikzpicture}
}
\caption{Application domains of synthetic data in human analysis.\label{fig:application-hierarchy}}
\end{figure*}

Motivated by the above, synthetic data has enabled a number of applications, listed in Table \ref{tab:synth-vs-real} and elaborated on in Section \ref{sec:usage}. 

Further, Figure \ref{fig:application-hierarchy} summarises application scenarios derived from the forthcoming literature survey.

\subsection{Benefits of synthetic data}
\label{sec:benefits}

Synthetic data can impart a performance boost to human analysis models, augment controllability and scalability, and mitigate privacy concerns. We here outline such benefits, whereas Section \ref{sec:usage} revisits relevant works.

\textit{Performance boost.} One ample application of synthetic data has been towards boosting the performance of human analysis models. Table \ref{tab:synth-vs-real} demonstrates such boost %the usefulness of synthetic data for boosting the performance of human analysis models in many different domains and over several metrics 
 by comparing the associated performance before and after the use of synthetic data in several domains such as  %More precisely, it documents the performance of various works in human analysis, which have reported their results based on synthetic and real data, respectively. It can be observed that many of the usage types introduced in the last subsection were applied to improve the performance across many application domains, such as 
 action recognition, crowd counting, face recognition, pose estimation, and gender classification. Moreover, Table \ref{tab:synth-vs-real} shows that synthetic evaluation datasets, including controlled labels, are exploited to evaluate the performance of %conduct performance analysis of 
 new algorithms and pre-trained models. In human analysis, the high fidelity of evaluation datasets has been mainly fuelled by the remarkable progress in the domain of conditional image synthesis, which enables the generation of \textit{synthetic mated samples} by manipulating single image semantics.  

\textit{Controllability and scalability.} The advances in generative models have enabled the generation of synthetic data, incorporating fine-grained control over semantics. Consequently, synthetic datasets can be created to balance important factors of variation (\textit{e.g.}, the proportion of images pertained to male and female subjects), reducing biases caused by the unequal class distributions often observed in real-world datasets. Further, the employment of image synthesis models enables the generation of large-scale synthetic datasets, a factor known to correlate with the performance of DNNs. 
 
\textit{Mitigating privacy concerns.} Finally, fully-synthetic datasets reduce privacy concerns related to the distribution and processing of sensitive human data. Despite known incidents of information leaks of GANs \cite{feng2021gans}\cite{tinsley2021face}\cite{fredrikson2015model}, the reconstruction of training samples remains a challenge, as opposed to real data processing. We note that such \textit{information leakage} is of concern and a set of related countermeasures have been identified, such as the concepts of \textit{differential privacy} \cite{xu2019ganobfuscator} and \textit{precision reduction} \cite{fredrikson2015model}. Due to legal and privacy concerns, large-scale biometric datasets, such as MegaFace~\cite{kemelmacher2016megaface}, have been withdrawn from public channels. Instead, we envision that large-scale synthetic datasets will be made publicly available for DNN training and evaluation.

\subsection{Human analysis}

This survey defines \textit{human analysis} as the analysis of human characteristics, behaviour, and interaction with the physical world. Such analysis has myriad applications, summarised in Figure \ref{fig:application-hierarchy}. To elaborate, we note the following applications. % provides an overview of the fields of interest, which are further described as follows: 

\begin{itemize}

    \item \textbf{Action recognition} focuses on recognizing  activity of individual(s) from a series of observations from data subjects and their environment~\cite{roitberg2021let}.
    
    \item \textbf{Biometric recognition} refers to the automated recognition of individuals based on their biological and behavioural characteristics \cite{ISO-IEC-2382-37-170206}.
    
    \item \textbf{Emotion recognition} refers to the process of classifying human emotion~\cite{bozorgtabar2019using}.
    
    \item \textbf{Soft biometric classification} aims at automated classification of human characteristics in pre-defined categories, such as demographic, anthropometric or behavioural groups \cite{7273870}.  
    
    \item \textbf{Presentation attack detection} (PAD) refers to the automated determination of a presentation to the biometric data capture subsystem to interfere with the operation of the biometric system \cite{ISO-IEC-30107-1-PAD-Framework-160115}.
    
    \item \textbf{Human interaction recognition} is the task of analysing human interactions of at least two individuals who are interrelated to each other (\textit{e.g.}, handshaking)~\cite{shu2019hierarchical}. 
    
    \item \textbf{People detection/counting} denotes the detection or counting of individuals within a given image or video~\cite{aranjuelo2021key}\cite{wang2021pixel}.

    \item \textbf{Semantic segmentation} signifies the pixel-based image classification with the goal of tracking human bodies~\cite{varol2017learning} or body parts~\cite{wood2021fake} in a given image or video.
    
    \item \textbf{Pose estimation} estimates the transformation of the human body~\cite{menier20063d} or head~\cite{hempel20226d} from a reference pose, given an image or a 3D scan.

    \item \textbf{Person Re-Identification} is the task of identifying an individual captured in images and videos acquired from different cameras or camera angles~\cite{fu2021unsupervised}~\cite{sun2019dissecting}
    
    \item \textbf{Anomaly detection} refers to classifiers trained to detect human behaviours, interactions, or movements deviating from normality~\cite{kolberg2021anomaly}.
    
     \item \textbf {Medical analysis} refers to the automated analysis of data collected in medical applications with the greater goal of restoring and maintaining human health. In this survey, synthetic data in medical applications is considered out-of-scope, and interested readers are referred to the work of Chen \textit{et al.}~\cite{chen2021synthetic}.

\end{itemize}

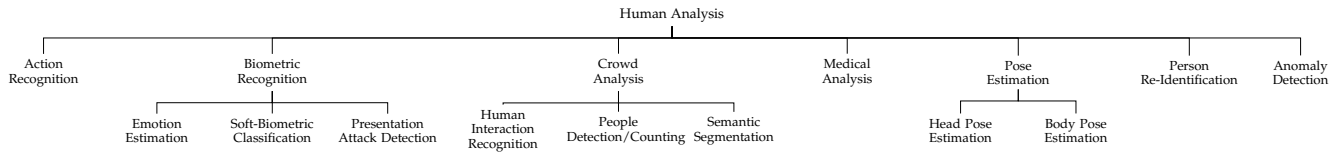
\begin {figure*}%[!hbtp]
\resizebox{\linewidth}{!}{%
\centering
\begin{tikzpicture}[level distance=40pt]
  \tikzset{edge from parent/.style={draw,-, fill=none,
    edge from parent path={(\tikzparentnode.south) -- +(0,-8pt) -| (\tikzchildnode)}}}
  \tikzset{every tree node/.style={align=center, fill=none}}
  \tikzset{every level 1 node/.style={font=\small, text width=2.4cm}}
  \tikzset{every level 2 node/.style={font=\small, text width=2.5cm}}

  \Tree [.{Human Analysis}
    [.{Action \\ Recognition} ]
    [.{Biometric \\ Recognition} [.{Emotion \\ Estimation} ] [.{Soft-Biometric \\ Classification} ] [.{Presentation  \\ Attack Detection} ]]
    [.{Crowd \\ Analysis} [.{Human Interaction \\ Recognition} ] [.{People \\ Detection/Counting} ] [.{Semantic \\ Segmentation} ]]
    [.{Medical \\ Analysis} ]
    [.{Pose \\ Estimation} [.{Head Pose \\ Estimation} ] [.{Body Pose \\ Estimation} ]]
    [.{Person \\ Re-Identification} ]
    [.{Anomaly \\ Detection} ]
    ]
\end{tikzpicture}
}
\caption{Application domains in human analysis\label{fig:application-hierarchy}}
\end{figure*}

\begin{table*}
\caption{Performance of human analysis models trained with and w/o synthetic data. Numbers given in \% (DE=data enrichment, DR=data replacement, EER=equal error rate, MAE=mean absolute error, MSE=mean square error, FNMR=false non-match rate, U=Illumination, E=Expression, P=Pose).}
\label{tab:synth-vs-real}
\begin{center}
\resizebox{\linewidth}{!}{%
\begin{tabular}{|l||l|l||l||c|c|c|}
\hline
\textbf{Reference} & \textbf{Application Domain} & \textbf{Application Type} & \textbf{Metric} & \textbf{w/o synthetic data} & \textbf{DE} & \textbf{DR}  \\  \hline\hline

Aranjuelo \textit{et al.}~\cite{aranjuelo2021key}  & people detection & Augmented Training & Average Precision ($\uparrow$) & 70 & 82 & - \\ \hline

Wang \textit{et al.}~\cite{wang2021pixel}  & crowd counting & Synthetic Training & MAE ($\downarrow$) & $275.5$ & - & $225.9$ \\ \hline

Yadav \textit{et al.}~\cite{yadav2019synthesizing}& unseen iris PAD & Augmented Training & EER ($\downarrow$) & $25.18$ & $18.52$ & - \\ \hline

Grosz and Jain~\cite{grosz2022spoofgan}& Fingerprint PAD & Augmented and Synthetic Training & Accuracy ($\downarrow$) & $99.52$ & 100 & $36.53$ \\ \hline

Bird \textit{et al.}~\cite{bird2020lstm}& speaker recognition & Model Initialization & Average Accuracy ($\uparrow$) & $95.48$ & $99.35$ & - \\ \hline

Tapia \textit{et al.}~\cite{tapia2019soft}& Gender classification from periocular images & Evaluation & Accuracy ($\uparrow$) & $82.76$ & - & $91.9$\\ \hline

Anton \textit{et al.}~\cite{anton2020modification}& Voice Recognition & Augmented Training & Accuracy ($\uparrow$) & $51.55$ & $57.76$ & - \\
\hline

Acien \textit{et al.}~\cite{acien2020becaptcha}& Bot Detection & Augmented Training & EER ($\downarrow$) & $23.2$ & $0.8$ &  - \\
\hline

Han \textit{et al.}~\cite{han2018improving}& Face Detection & Evaluation & Average Precision ($\uparrow$) & 64 & $74.5$ &  \\
\hline

Basak \textit{et al.}~\cite{basak2021learning}& head pose estimation & Domain Adaptation & MAE ($\downarrow$) & $6.34$ & - & $5.13$  \\
\hline

Bird \textit{et al.}~\cite{bird2020overcoming}& speaker recognition & Model Initialization & Accuracy ($\uparrow$) & $96.58$ & $98.83$ & \\ \hline

Dou \textit{et al.}~\cite{dou2021versatilegait}& Gait recognition & Augmented Training & Rank-1 Accuracy ($\uparrow$) & $95.0$ & $96.4$ &  - \\
\hline

Piplani \textit{et al.}~\cite{piplani2018faking}& passthought authentication & Augmented Training & Accuracy ($\uparrow$) & $90.8$ & 95 & - \\
\hline

Buriro \textit{et al.}~\cite{buriro2021swipegan}& Smartphone User Authentication & Augmented Training & true accept rate ($\uparrow$) & $84.66$ & $91.65$ & - \\
\hline

Buriro \textit{et al.}~\cite{buriro2021swipegan}& Smartphone User Authentication & Augmented Training & false accept rate ($\downarrow$) & $14.78$ & $11.04$ &  \\
\hline

Gouiaa  \textit{et al.}~\cite{gouiaa2017learning}& Posture recognition & Augmented Training & Accuracy ($\uparrow$) & $94.58$ & 99 & - \\
\hline

Ruiz \textit{et al.}~\cite{ruiz2020off}& Signature verification & Augmented Training & EER ($\downarrow$) & $11.11$ & $4.9$ & - \\ \hline

Kortylewski \textit{et al.}~\cite{kortylewski2018training}& Face recognition & Model Initialization & Accuracy ($\uparrow$) & $91.2$ & $93.3$ & -  \\
\hline

Trigueros \textit{et al.}~\cite{trigueros2018generating}& Face recognition & Augmented Training & Accuracy ($\uparrow$) & $67.58$ & $69.02$ &  - \\
\hline

Zhai \textit{et al.}~\cite{zhai2021demodalizing}& Face recognition & Augmented Training & True Accept Rate ($\uparrow$) & $61.72$ & $86.66$ & -  \\
\hline

Chen \textit{et al.}~\cite{chen2019emotionalgan}& Emotion State Classification & Augmented Training & Accuracy ($\uparrow$) & $58.6$ & $64.5$ & - \\
\hline

% No synth vs real comparison given in paper
% \cite{dou2017end}& Face Reconstruction & & RMSE ($\downarrow$) & 4.23 & 4.00  &  \\ \hline

Melo\textit{et al.}~\cite{melo2019deep}& Signature Recognition & Synthetic Training & EER ($\downarrow$) & $10.26$ & - &  $9.74$ \\ \hline

\"Oz \textit{et al.}~\cite{oz2021use}& Eye Segmentation & Augmented Training & mIoU ($\uparrow$) & 73 & $75.4$ &  - \\
\hline

Wang \textit{et al.}~\cite{wang2019learning}& Crowd Counting & Model Initialization & MSE ($\downarrow$) & $14.3$ & 13  & - \\ \hline
 
Irtem \textit{et al.}~\cite{irtem2019impact}& Fingerprint Classification & Joint and Synthetic Training & Classification accuracy ($\uparrow$) & $91.9$ & $95.53$ &  $69.47$   \\ \hline

Engelsma \textit{et al.} ~\cite{engelsma2022printsgan}& Fingerprint Recognition & Model Initialization & True acceptance rate ($\uparrow$) & $73.37$ & $87.03$ & -  \\ \hline

Bozorgtabar \textit{et al.}~\cite{bozorgtabar2019using}& Expression Recognition & Domain Adaptation & accuracy ($\uparrow$) & $70.15$ & -  & $72.1$ \\ \hline

Qiu \textit{et al.}~\cite{qiu2021synface}& Face Recognition & Augmented and Synthetic Training & accuracy ($\uparrow$) & $91.22$ & $95.78$  & $91.97$  \\
\hline

Kortylewski \textit{et al.}~\cite{kortylewski2019analyzing}& Face Recognition & Model Initialization  & accuracy ($\uparrow$) & $91.2$ & $93.3$ & $88.9$   \\ \hline

Colbois \textit{et al.}~\cite{colbois2021use}& Face Recognition & Evaluation & FNMR U/E/P ($\downarrow$) & 11/3/55 & -  & 12/25/51 \\ \hline

Marriott \textit{et al.}~\cite{marriott20213d} & Pose-invariant Face Recognition & Augmented Training & accuracy ($\uparrow$) & $93.59$ & $95.29$  & - \\ \hline

Wood \textit{et al.}~\cite{wood2021fake} & Face Segmentation & Synthetic Training & $F_1$ score ($\uparrow$) & $91.6$ & -  & 92 \\ \hline

Ahmed \textit{et al.}~\cite{ahmed2019facial} & Face Expression Classification & Augmented Training & Accuracy ($\uparrow$) & $92.95$ & $96.24$  & - \\ \hline

Niinuma \textit{et al.}~\cite{niinuma2021synthetic} & Face Expression Classification & Synthetic Training & Inter-rater reliability ($\uparrow$) & $48.9$ & -  & $52.5$ \\ \hline

Abbasnejad \textit{et al.}~\cite{abbasnejad2017using} & Face Expression Classification & Augmented Training & $F_1$ score ($\uparrow$) & $60.52$ & $86.59$  & - \\ \hline

Varol \textit{et al.}~\cite{varol2021synthetic} & Action Recognition & Augmented Training & Accuracy $0^{\circ}/45^{\circ}/90^{\circ}$ ($\uparrow$) & $88.8$/$78.2$/$57.3$ & $90.5$/$83.3$/68  & - \\ \hline

Hatay \textit{et al.}~\cite{hatay2021learning} & (Phone) Action Recognition & Model Initialization & Accuracy ($\uparrow$) & $95.83$ & $96.67$ & - \\ \hline

Souza  \textit{et al.}~\cite{de2017procedural} & Action Recognition & Augmented Training & Accuracy ($\uparrow$) & $93.3$ & $92.7$ & - \\ \hline

Varol \textit{et al.}~\cite{varol2017learning} & Human Body Segmentation & Model Initialization & Accuracy ($\uparrow$) & $58.54$ & $67.72$ & $56.51$ \\ \hline

Priesnitz \textit{et al.}~\cite{priesnitz2022syncolfinger} & Contactless Fingerprint Recognition & Evaluation & Avg. EER ($\downarrow$) & $30.93$ & - & $3.55$  \\ \hline

\end{tabular}
}
\end{center}
\end{table*}

% You must have at least 2 lines in the paragraph with the drop letter
% (should never be an issue)
\section{How can synthetic datasets be generated?}
\label{sec:generation}
%Thanks to the benefits offered by synthetic data, the generation of synthetic datasets for human analysis has intrigued researchers for a long time. The 
Initial approaches for synthetic data generation generally exploit \textit{mathematical modelling}, \textit{3D rendering tools} or \textit{perturbations using classical and hand-crafted} means. However, the success of deep neural networks in image generation has catapulted \textit{dynamic perturbations} and \textit{deep neural networks} as primary generation models. We proceed to provide details on such synthetic data generation methods.% are provided next.

 \begin{figure}
		\centering \includegraphics[scale=0.6]{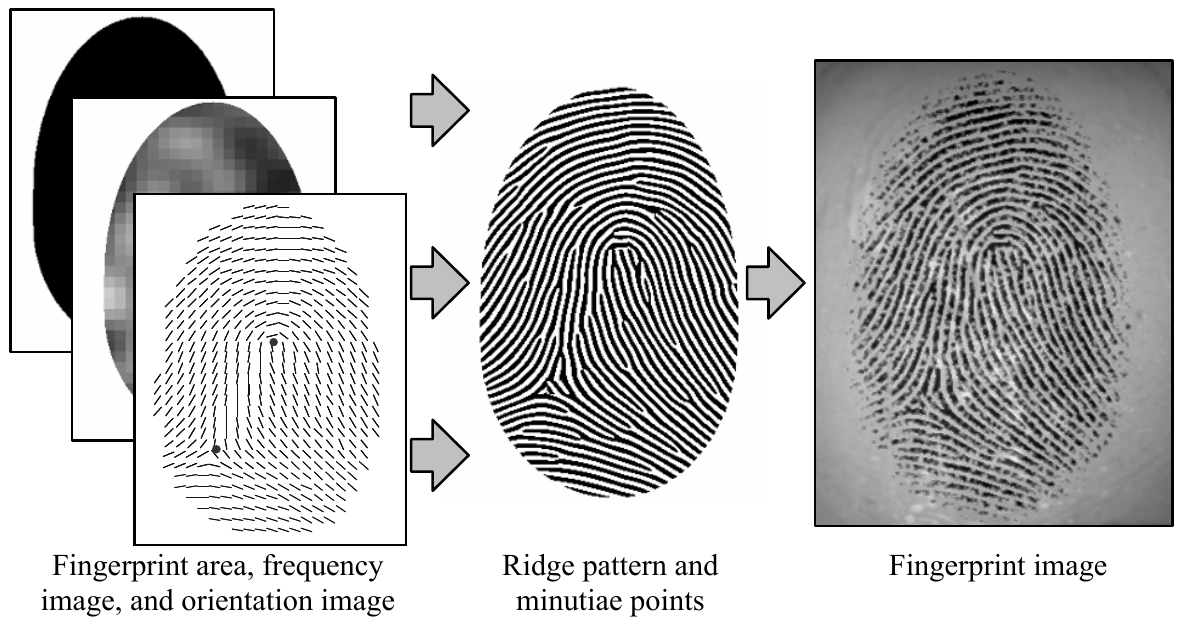}
		\caption{Cappelli \textit{et al.} \cite{maltoni2009synthetic} mathematically modelled the distribution of fingerprints and proposed SFinGe to generate synthetic fingerprints.}
		\label{mathematical_modeling}
\end{figure}
\subsection{Mathematical modelling}
\label{sec:mathematical_modeling}
Mathematical modelling constitutes an early approach for generating human data aimed at approximating the distribution of real human data through mathematical modelling. Sampling from the approximated model can then be used to generate synthetic samples and exploit them in downstream human analysis tasks. Approximation of the mathematical model pertaining to the human data requires domain expertise and a careful understanding of model parameters. A popular mathematical modelling-based synthetic fingerprint generation (SFinGe) is proposed by Cappelli \textit{et al.} \cite{maltoni2009synthetic} (see Figure \ref{mathematical_modeling}). The authors exploit domain expertise to define a fingerprint orientation model characterized by the number and location of the fingerprint cores and deltas. The synthetic fingerprint generation starts from initializing the locations of core and deltas, followed by ridge orientation and density generation. Subsequently, the authors apply space-invariant linear filtering to obtain a binarized good quality fingerprint image. Lastly, domain-specific noise is introduced to simulate realistic greyscale fingerprint images. Approaches exploiting mathematical modelling using domain knowledge for synthetic data generation include handwriting recognition \cite{carmona2017temporal,ferrer2013synthetic, o2009development,ferrer2014static}, finger vein recognition \cite{mciHillerstroem2014}, hand shape recognition \cite{cootes1995active}, face recognition \cite{osadchy2017g}, keystroke recognition \cite{migdal2019statistical} and iris recognition \cite{drozdowski2017sic,friedman2017synthetic}.

\subsection{3D rendering tools} 
\begin{figure}
		\centering \includegraphics[scale=1.15]{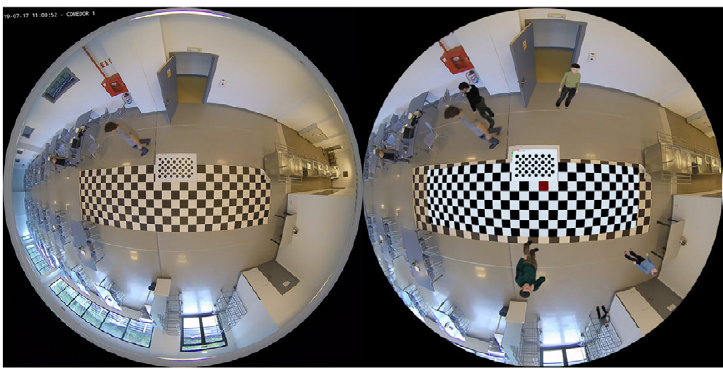}
		\caption{Aranjuelo \textit{et al.} \cite{aranjuelo2021key} utilized 3ds Max software to virtually render humans (right) on a real scene (left) for detection of moving subjects.}
		\label{synthetic_3d_example1}
\end{figure}
Several studies exploit 3D modelling to create mathematical representations of the three-dimensional surface of the object of interest. Subsequently, a 3D rendering tool is exploited to render images corresponding to a 3D model. Han \textit{et al.} \cite{han2018improving} argued that the generation of synthetic samples in 3D space allows for the incorporation of extreme changes in illumination, viewpoint, occlusion, scale, and background. Additionally, rendering engines allow precise control over environmental conditions such as pose variations, lighting, and object geometry leading to accurate annotations, which are often acquired for a real dataset. Most popular 3D rendering tools include Blender\footnote{\url{https://www.blender.org/}}, Maya\footnote{\url{https://www.autodesk.fr/products/maya/overview}}, 3ds Max\footnote{\url{https://www.autodesk.com/products/3ds-max/overview}}, Cinema 4D\footnote{\url{https://www.maxon.net/en/cinema-4d}}, Unity\footnote{\url{https://www.unity3D.com}}, and Unreal Engine\footnote{\url{https://www.unrealengine.com}}.

\par
\begin{figure}
		\centering \includegraphics[scale=0.75]{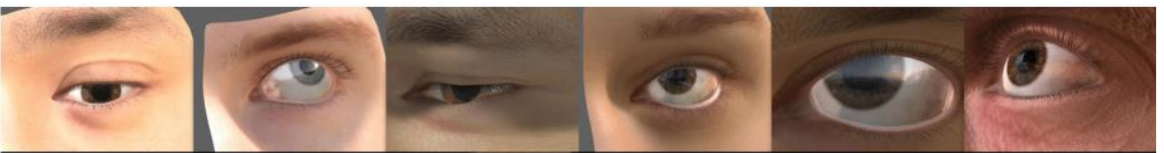}
		\caption{Öz \textit{et al.} \cite{oz2021use} generated synthetic eye images employing UnityEyes \cite{wood2016learning}, a 3D rendering tool. The synthetic data is used for learning eye region segmentation.}
		\label{synthetic_3d_example2}
\end{figure}

 Aranjuelo \textit{et al.} \cite{aranjuelo2021key} virtually rendered humans on real scenes for application in detection of individuals (see Figure \ref{synthetic_3d_example1}). Similarly, Öz \textit{et al.} \cite{oz2021use} used a 3D rendering tool to generate synthetic eye images and exploit the generated samples to learn eye region segmentation (see Figure \ref{synthetic_3d_example2}). Other studies exploiting 3D rendering tools for generating synthetic data spanned applications in re-identification of individuals \cite{wan2020person}, face recognition \cite{feng2015cascaded} \cite{liu20193d} \cite{han2018improving} and gait recognition \cite{charalambous2016data}.

% Blender, Maya, 3ds Max, Cinema 4D,
% Unity, and Unreal Engine are the most popularly used 3d modelling frameworks. 

\subsection{Input perturbations}
Perturbation of a given input is widely used to generate synthetic data. Input perturbation imparts either the introduction of noise through classical and hand-crafted methods or a learning-based approach. We proceed to provide a brief discussion on both approach types. 

\begin{figure}
		\centering \includegraphics[scale=0.65]{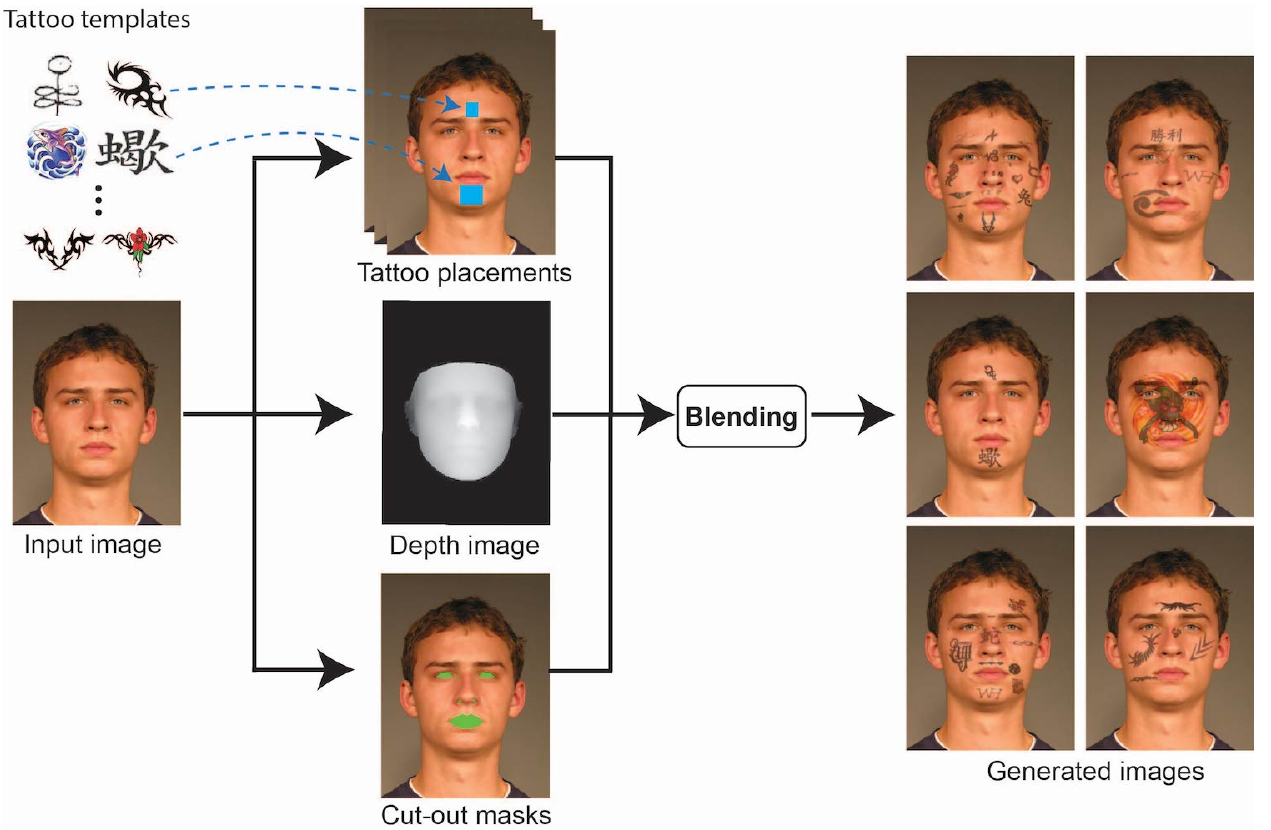}
		\caption{Ibsen \textit{et al.} \cite{ibsen2022face} generated synthetic face images with tattoos to learn tattoo removal from faces. Image processing operations were used to blend tattoos on faces. Further, operations such as colour-adjustment and Gaussian blurring were applied to increase realism in the synthetically tattooed faces.}
		\label{image_processing_perturbation_example2}
\end{figure}

\begin{figure}
		\centering \includegraphics[scale=0.6]{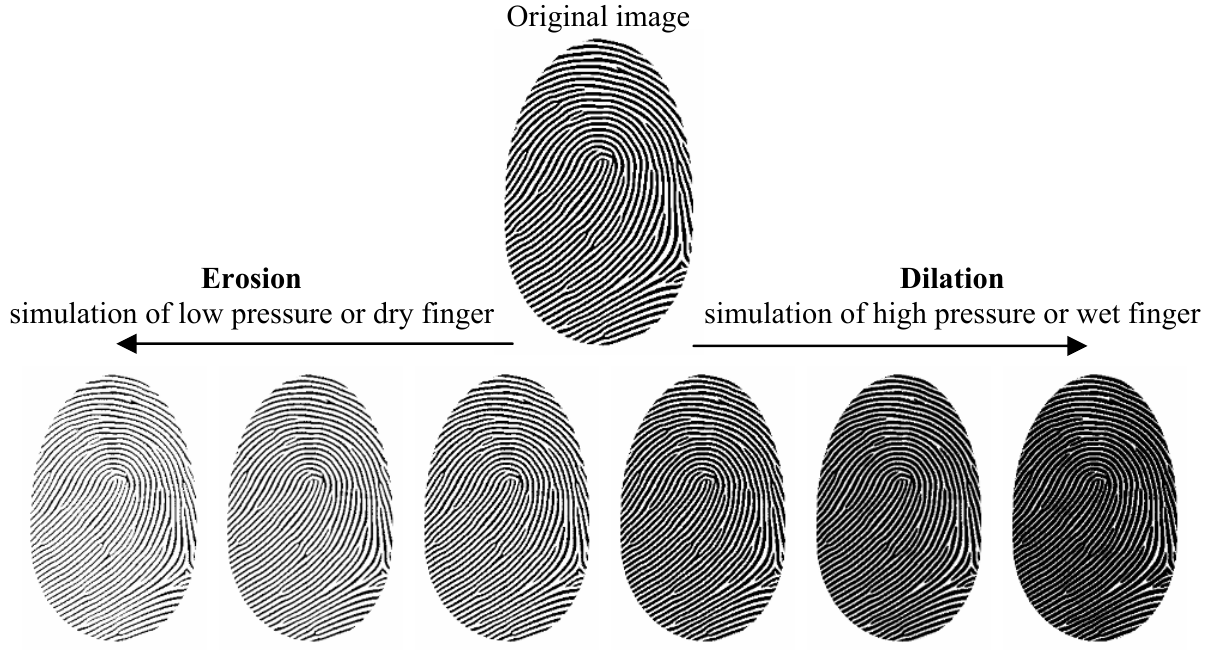}
		\caption{Cappelli \textit{et al.} \cite{maltoni2009synthetic} exploited morphological operations such as erosion and dilation to vary ridge thickness while generating multiple impressions of a fingerprint image.}
		\label{image_processing_perturbation_example1}
\end{figure}

\subsubsection{Perturbations using classical and hand-crafted methods} 
Classical and hand-crafted methods can perturb a given input to either introduce variations in the available data or simulate cases that are difficult to capture otherwise. Most prominent classical and hand-crafted methods include Gaussian blurring, image blending, colour jittering, horizontal and vertical flipping, rotation, translation, as well as affine transformations. Some studies utilize morphological operations such as erosion and dilation to generate synthetic data samples. Following this direction of synthetic data generation, Ibsen \textit{et al.} \cite{ibsen2022face} exploited image processing techniques to synthetically blend tattoos on human faces (see Figure \ref{image_processing_perturbation_example2}). Similarly, Cappelli \textit{et al.} \cite{maltoni2009synthetic} generated synthetic multiple impressions from a given input fingerprint using morphological operations (see Figure \ref{image_processing_perturbation_example1}). Other studies that generate synthetic data using classical methods have been instrumental in face recognition \cite{nojavanasghari2017hand2face} \cite{anton2020modification}, fingerprint recognition \cite{cao2015latent} \cite{joshi2021data} \cite{dieckmann2019fingerprint}, iris recognition \cite{cardoso2013iris} \cite{fuentes2019hybrid} and re-identification of individuals \cite{fu2021unsupervised} \cite{yin2022unsupervised}.

\begin{figure}
		\centering \includegraphics[scale=1]{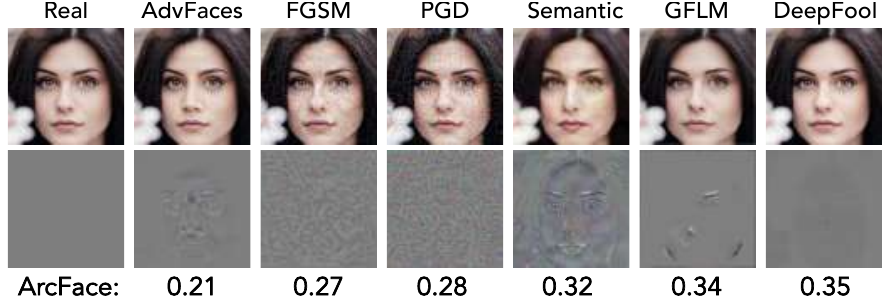}
		\caption{Jain \textit{et al.} \cite{jain2021biometrics} \cite{deb2020faceguard} dynamically perturb a face image using six different adversarial training mechanisms (top row). The corresponding perturbations are provided in the bottom row. The authors demonstrate that synthetic faces generated using dynamic perturbations can increase face comparison score (obtained using ArcFace) in non-mated comparison trials.}
		\label{dynamic_perturbation_example}
\end{figure}

\subsubsection{Dynamic perturbations}
A dynamic perturbation is defined as an input-specific perturbation introduced through an adversarial training mechanism such that a learning-based human analysis model is likely to make an erroneous prediction \cite{li2020automatic}. Training a human analysis model with the synthetic data generated using dynamic perturbations is beneficial for regularization and improvement of robustness. Following this approach, several studies generate synthetic data using adversarial training. Jain \textit{et al.} \cite{jain2021biometrics} \cite{deb2020faceguard} generated synthetic non-mated facial images using dynamic perturbations that obtain high comparison scores (see Figure \ref{dynamic_perturbation_example}). Other studies in human analysis exploiting dynamic perturbations include applications in speaker identification \cite{li2020automatic}, re-identification of individuals \cite{wang2020transferable}, face recognition \cite{zhong2020towards}, iris recognition \cite{soleymani2019adversarial} and fingerprint recognition \cite{marrone2021transferability}. 

% \cite{zhou2020uncertainty} \cite{wang2019adversarial}.

% survey: \cite{jain2021biometrics}

\subsection{Deep neural networks}
Deep neural networks (DNNs) represent state-of-the-art architectures for generating synthetic data for among others, applications in human analysis. By revisiting related literature, we identify following categories for doing so. %find that the deep neural networks used to generate synthetic data of humans can be broadly categorized as follows:

% \subsubsection{Convolutional Neural Networks}

\subsubsection{Recurrent neural networks}
\label{sec:DNNs}
A recurrent neural network (RNN) is a DNN designed to process time-series, as well as sequential or variable-length input data. Such models are designed for applications, where input data samples depend on the previous data samples, as RNNs are aimed at capturing dependencies between data samples. Towards capturing long-range dependencies, state-of-the-art RNNs exploit long short-term memory (LSTM) and gated recurrent units (GRU) \cite{mirsky2021creation} to store information from previous inputs or states and generate the subsequent output of the input sequence. An LSTM comprises three gates: input, output and forget gate, while a GRU incorporates a reset and an update gate. These gates determine the most informative part of the input to make a prediction in the future.

\par One of the applications exploiting RNN to generate synthetic data is the contribution of Bird \textit{et al.} \cite{bird2020overcoming}, where a character-level RNN is exploited to generate audio sentences for speaker identification. In addition, RNNs are employed for generating deep fakes, where these architecture render continuous realistic flow in audio or video \cite{mirsky2021creation}.

%  Melo \textit{et al.} \cite{melo2019deep} propose a fully convolution network to translate online signatures to synthetic offline signatures.
 
%  \subsubsection{Encoder-Decoder Models}

\begin{figure}
		\centering \includegraphics[width=0.45\textwidth]{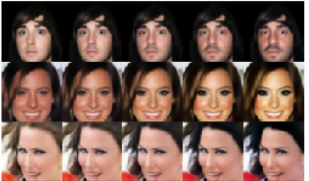}
		\caption{Mondal \textit{et al.} \cite{mondal2021flexae} proposed a variational auto-encoder model for generating synthetic faces, controlling attributes such as big nose, makeup, black hair, smile and gender.}
		\label{synthetic_autoencoder_example}
\end{figure}
\subsubsection{Auto-Encoders}
 Auto-Encoder (AE) based generative models constitute a pair of encoder and decoder networks. While the encoder network learns an efficient representation of the input, the decoder network generates an output corresponding to the given latent vector provided as output by the encoder network. These models generate synthetic data by learning the joint distribution of the latent space and the training data. Such models are generally regularized by imposing a prior distribution on the latent space to facilitate generation during inference \cite{kingma2013auto}. Prominent auto-encoder architectures for synthetic data generation include variational auto-encoder \cite{kingma2013auto}, adversarial autoencoder \cite{makhzani2015adversarial} and Wasserstein auto-encoder \cite{tolstikhin2018wasserstein}, which includes a Gaussian prior. However, the Gaussian prior is simplistic and might fail to capture complex latent distributions. To alleviate this limitation, rich classes of distributional priors have been explored \cite{mondal2021flexae, tomczak2018vae, dai2018diagnosing}. Several research efforts have attempted to learn disentangled representations in the latent space of the VAE \cite{higgins2016beta, kim2018disentangling} (see Figure \ref{synthetic_autoencoder_example}). Such a factored representation is beneficial in interpolating the latent space leading to the generation of diverse samples and plausible modification in input data. Despite offering interpretable inference, stable training, and an efficient sampling procedure, the generation quality of VAEs is not as impressive as the one achieved by GANs \cite{dai2018diagnosing, pmlr-v124-mondal20a}. Next, we discuss the most widely used state-of-the-art deep generative framework, namely GANs \cite{goodfellow2014generative}.
 
 \begin{figure}
		\centering \includegraphics[width=0.48\textwidth]{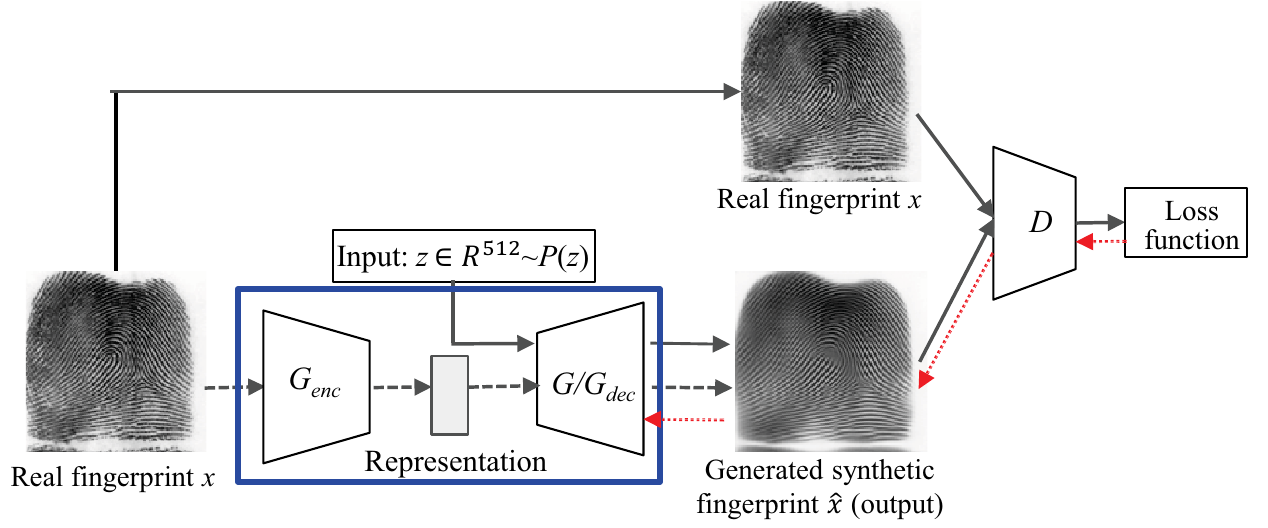}
		\caption{Cao and Jain \cite{cao2018fingerprint} generated synthetic fingerprints using improved Wasserstein GAN that was trained to upscale a random noise vector into a fingerprint image.}
		\label{synthetic_gan_example1}
\end{figure}
\begin{figure}
		\centering \includegraphics[width=0.48\textwidth]{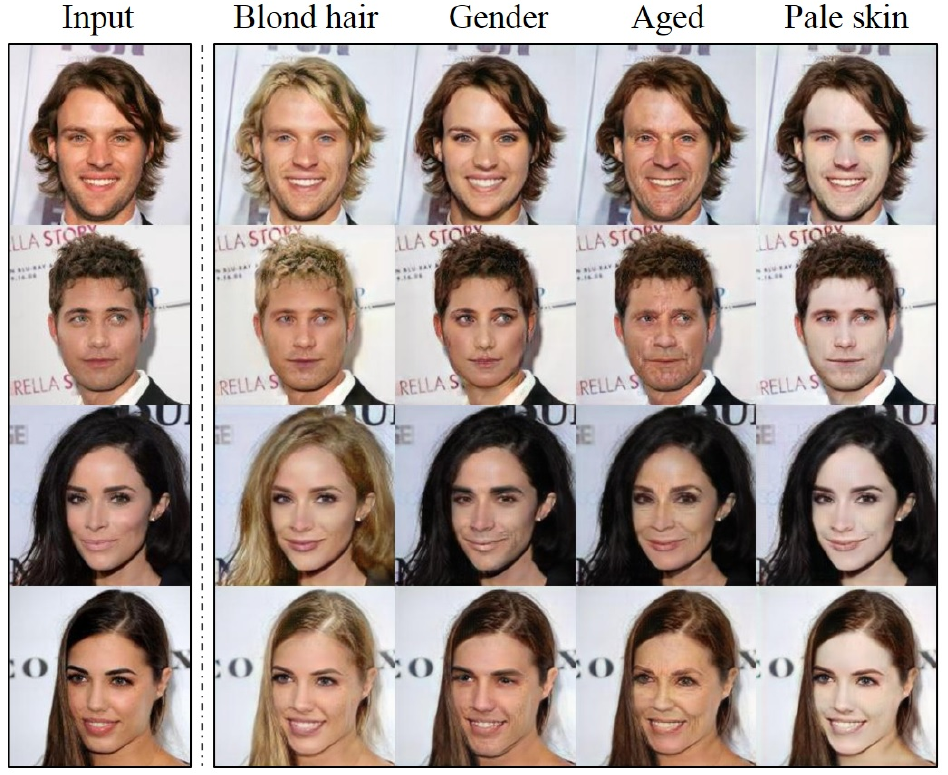}
		\caption{Choi \textit{et al.} \cite{choi2018stargan} proposed StarGAN, a generative adversarial network that was able to alter attributes of a given face image. The generated synthetic faces have been commonly used as deepfakes.}
		\label{synthetic_gan_example2}
\end{figure}
% \begin{enumerate}
%     \item {Noise to Image Generation:} 
 % \item {Image to Image Translation:}
    % \begin{itemize}
    %     \item {Pix2pix:}
        
    %      \item {Cycle-GAN:}
         
    %      \item {StyleGAN:}
    % \end{itemize}
% \end{enumerate}
    
\subsubsection{Generative adversarial networks (GANs)}
Goodfellow \textit{et al.} in their seminal work \cite{goodfellow2014generative} proposed a framework incorporating two networks, a generator and a discriminator. The generator is learns distribution of training samples, whereas the discriminator network is aimed at classifying whether the input samples stem from the training set or are generated by the generator (real or fake). Both networks are trained in an adversarial manner (zero-sum game), and the framework targets to facilitate improved approximation of true distribution by the generative model \cite{mondal2020c}. Hence, the name \textit{generative adversarial network}. GANs are broadly categorized as \textit{noise to image translation GANs} or \textit{image to image translation GANs}. Noise to image translation GANs are trained to upscale a randomly sampled noise vector to a realistic image, whereas the image to image translation GANs are trained to transform a given image to another image.

% Generative adversarial networks (GANs) are the state-of-the-art architectures for generating synthetic data \cite{goodfellow2014generative} \cite{mondal2020c}. A GAN constitutes of a generator and a discriminator network. The success of a GAN is attributed to the existence of a discriminator network that classifies the input sample as real or fake. Thus, ensuring generation of realistic synthetic samples. 
Prominent noise to image translation GANs include DCGAN \cite{radford2015unsupervised} and Wasserstein GAN \cite{arjovsky2017wasserstein}, whereas frequently empolyed image to image translation GANs include pix2pix \cite{isola2017image} and Cycle-GAN \cite{zhu2017unpaired}. Several studies in human exploited GANs to generate synthetic data  \cite{chen2021joint} \cite{tapia2019soft} \cite{acien2020becaptcha} \cite{buriro2021swipegan} \cite{piplani2018faking} \cite{trigueros2018generating} \cite{zhai2021demodalizing} \cite{bozorgtabar2019using} \cite{li2020gait}. One such study includes the contribution of Cao and Jain \cite{cao2018fingerprint}. The authors generated synthetic fingerprints using noise to image translation GAN (see Figure \ref{synthetic_gan_example1}). Similarly, Choi \textit{etal.} \cite{choi2018stargan} proposed an image to image translation GAN to modify attributes in facial images (see Figure \ref{synthetic_gan_example2}).

\section{How can synthetic data be utilized?}
\label{sec:usage}
%Motivated by time and resource constraints, as well as legal difficulties in the collection and release of human data, 
Synthetic data is frequently used to \textit{simulate complex scenarios} for which the data collection is particularly challenging, \textit{overcome privacy issues} observed for collection of real human analysis datasets, \textit{increase the size and diversity of training datasets}, as well as \textit{mitigate bias} in real training datasets. Furthermore, looking at the challenge in collecting large-scale datasets, synthetic data is widely used in \textit{scalability analysis} of systems. Additionally, as obtaining annotations can be both time-consuming and expensive, \textit{synthetic data, whose annotations can be automatically derived} is popularly used. With \textit{consistency regularization} techniques, synthetic data is used to learn generalizable models. Synthetic data can also be employed to produce \textit{presentation attacks} on human authentication systems. We proceed to provide details on different usage of synthetic data.

%\begin{enumerate}
 
 \begin{figure}
		\centering \includegraphics[scale=1]{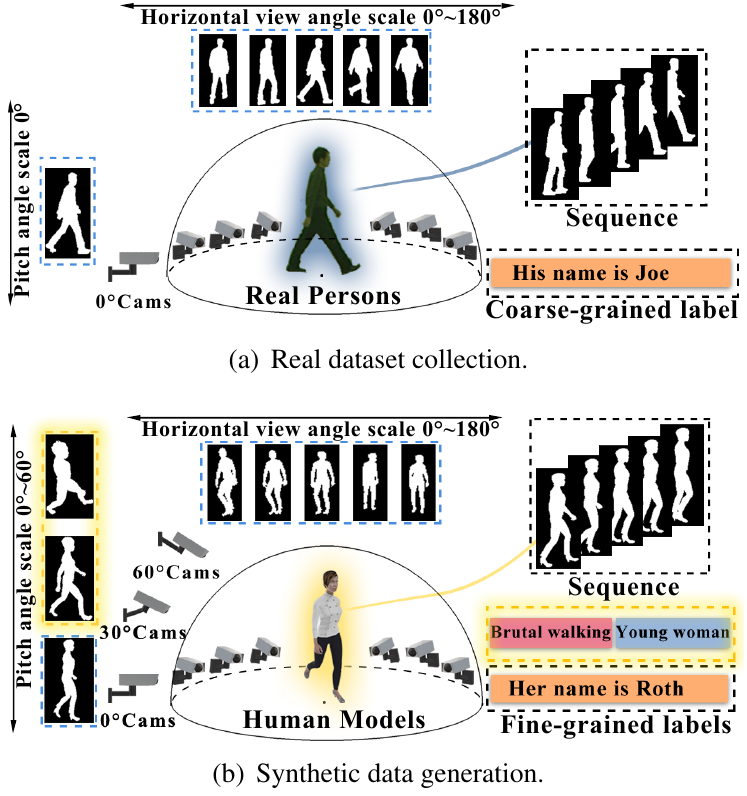}
		\caption{Dou \textit{et al.} \cite{dou2021versatilegait} discussed the limitations of existing real databases for video-based gait recognition in capturing complex scenarios. For instance, the authors discussed that (a) real datasets are only acquired with a single camera pitch angle and (b) on the other hand, the synthetic dataset is generated with a diverse range of camera pitch angles. Thus, synthetic data can be used to simulate complex scenarios, which are otherwise difficult to acquire for a real dataset for human analysis.}
		\label{complex_scenarios}
\end{figure}

    \subsection{Simulating complex scenarios} 
    Dou \textit{et al.} \cite{dou2021versatilegait} argued that existing real databases for video-based gait recognition do not possess examples of complicated scenarios that can be crucial to obtaining satisfactory performance in real-world applications. For instance, real datasets are captured under ideal settings with only a single camera pitch angle (see Figure \ref{complex_scenarios}). %The authors also discuss the limitation of the existing databases, such as t
    Specifically in the OU-MVLP dataset \cite{takemura2018multi} for gait recognition % and point out that in this standard dataset for gait recognition, the 
    subjects only walk twice without the change of bag or clothing, %. Furthermore, the authors also highlight that the existing real gait recognition databases contain 
    with only one subject appearing per video frame. However, real-world scenarios naturally include multiple walking individuals. %close to each other is a common real-world scenario ignored so far. 
    %Therefore, the authors concluded that the existing databases do not contain adequate examples of complex real-world scenarios. 
    Towards bridging this gap, the authors generated approximately one million synthetic silhouette sequences of 11,000 subjects. The resulting synthetic dataset VersatileGait comprises of gait sequences with a diverse range of camera pitch angles and fine-grained annotations of attributes. Furthermore, to promote the design of multi-person gait recognition algorithms, the authors also generated multi-person walking scenarios with up to three people walking simultaneously.
    
    \par Similarly, Aranjuelo \textit{et al.} \cite{aranjuelo2021key} argued that existing real datasets for human detection do not exploit omnidirectional cameras to capture a 360\textdegree ~view in surveillance videos. To take advantage of the 360\textdegree ~view, the authors proposed the subject detection model to be trained with synthetic data. % instead to train a subject detection model. 
    Other applications, exploiting synthetic data to simulate complex scenarios include the contributions of Lai \textit{et al.} \cite{lai2020synsig2vec} for generating synthetic skilled forgery attacks, Tabassi \textit{et al.} \cite{tabassi2018altered} for simulating altered fingerprints and the contributions of Arifoglu and Bouchachia \cite{arifoglu2019abnormal} for the simulation of (abnormal) behaviour observed for dementia patients.
    % Another recent example that exploits synthetic data for simulate complex scenarios includes the work of Ibsen \textit{et al.} \cite{ibsen2022face} that generate synthetically tattooed faces to learn tattoo removal from faces.
    
    % \par Similarly, Lai \textit{et al.} \cite{lai2020synsig2vec} highlight the difficulty in acquiring data pertaining to skilled forgery attacks on signature verification systems. To mitigate this challenge, the authors generate synthetic signatures by introducing distortions in real template signatures. The authors demonstrate that the model trained using only the synthetically attacked signatures outperforms existing state-of-the-art signature verification systems trained on real attacked signatures by 8.06\% on MCYT-100 database \cite{ortega2003mcyt}. Likewise, Arifoglu and Bouchachia \cite{arifoglu2019abnormal} generate synthetic data to simulate (abnormal) behaviour observed for dementia patients. 
    
    % Abnormal behaviour detection model is pre-trained on real dataset of healthy individuals. Unlabelled synthetic dataset of abnormal behaviours is used for adapting the model on abnormal behaviour cases.
    
    \begin{figure}
		\centering \includegraphics[scale=1]{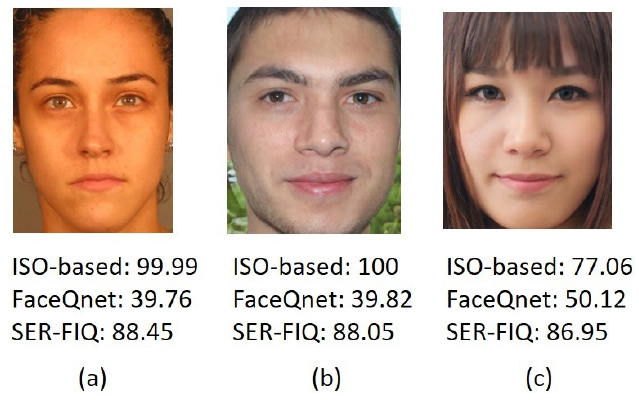}
		\caption{A comparison of (a) real face image (from FRGC-V2 face database \cite{phillips2005overview}) and its associated quality score compared to synthetic face images generated by Zhang \textit{et al.} \cite{zhang2021applicability} using (b)  StyleGAN \cite{karras2019style} and (c) StyleGAN2 \cite{karras2020analyzing}. The synthetic faces have similar quality scores to those obtained for a real face. The authors proposed exploiting synthetic images to assess the performance of face recognition systems. This is instrumental in addressing and responding to privacy constraints in sharing datasets.}
		\label{privacy_preserving}
\end{figure} 
  
  \begin{figure}
		\centering \includegraphics[scale=1]{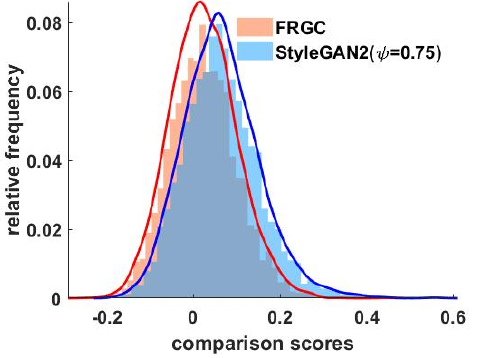}
		\caption{Zhang \textit{et al.} \cite{zhang2021applicability} compared the distribution of non-mated comparison scores of real (FRGC-V2 face database \cite{phillips2005overview}) versus synthetic faces generated. Only minor differences in non-mated comparison scores corresponding to synthetic data were observed, as seen in real data. These results illustrated the potential of synthetic data being utilized instead of real data, alleviating privacy issues.}
		\label{scores_synthetic_real_privacy}
\end{figure}
\subsection{Addressing privacy concerns} 
Data collection is often governed by strict rules to preserve the identity of individuals. For settings, in which data collection is challenging, generated synthetic data and perform experiments on such synthetic data \cite{carmona2017temporal}. However, a challenge with these applications has been to ensure that synthetic data has a similar distribution (for instance, distribution of minutiae in fingerprints \cite{cao2018fingerprint}, or distribution of sample quality scores \cite{cao2018fingerprint, zhang2021applicability}) as the real data. Many studies demonstrated that synthetic data with similar characteristics to the real data can be generated and used, rather than the privacy-constrained real data. One such study includes generation of $50,000$ synthetic face images each using StyleGAN \cite{karras2019style} and StyleGAN2 \cite{karras2020analyzing} for face recognition applications in face recognition systems at the Schengen border \cite{zhang2021applicability}. The authors demonstrated that realistic face images with image quality scores %(obtained using standard metrics for face image quality assessment)  
similar to real faces can be generated (see Figure \ref{privacy_preserving}). In addition, the authors compared face recognition performance of models trained on synthetic and real data and reported only minor differences, see Figure \ref{scores_synthetic_real_privacy}. Similar to Zhang \textit{et al.} \cite{zhang2021applicability},  Bozkir \textit{et al.} \cite{bozkir2020privacy} and Hillerström \textit{et al.} \cite{hillerstrom2014generation} proposed to generate synthetic data for applications implying gaze estimation and finger vein recognition, respectively, in order to circumvent privacy issues, occurring when publicly sharing human data.
    
%   Bozkir \textit{et al.} \cite{bozkir2020privacy} use synthetic images from UnityEyes \cite{wood2016learning} database to train a support vector regression model for gaze estimation. Hillerström and Kumar \cite{hillerstrom2014generation} propose to overcome privacy restrictions in finger vein recognition by generating synthetic finger vein samples.

  \begin{figure}
		\centering \includegraphics[scale=0.98]{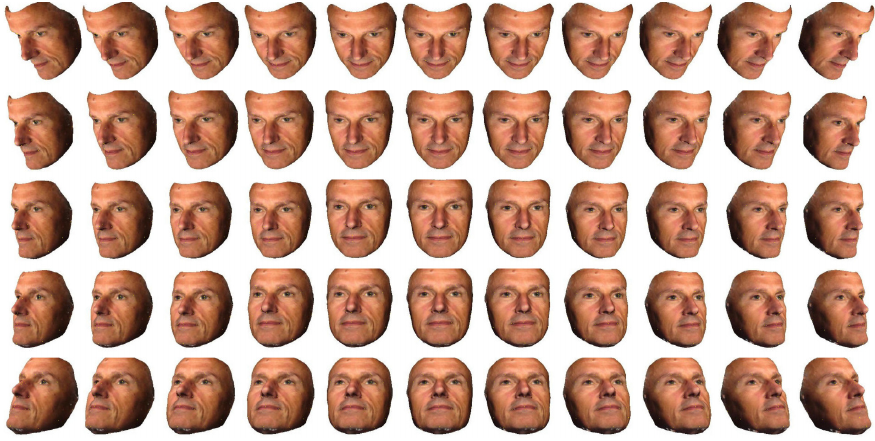}
		\caption{Feng \textit{et al.} \cite{feng2015cascaded} generated synthetic faces with 11 yaw rotations (ranging from \textminus50\textdegree~to 50\textdegree~over a step size of 5\textdegree) and 5 pitch rotations (ranging from \textminus30\textdegree~to 30\textdegree~over a step size of 5\textdegree). Therefore, augmenting the synthetic dataset with the real training set increased the size of the training set. Furthermore, synthetic data provided face images with diverse pose variations. As a result, facial landmark detection performance improved \textit{w.r.t.} variations in facial poses.}
\label{augmentation_synthetic}
\end{figure} 
 
  \begin{figure}
		\centering \includegraphics[scale=0.95]{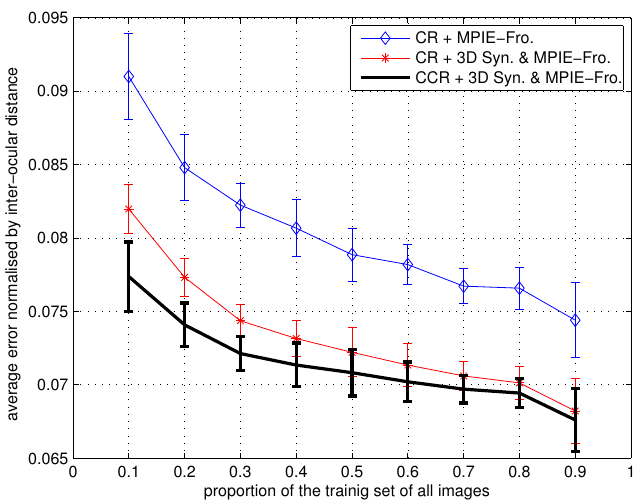}
		\caption{Feng \textit{et al.} \cite{feng2015cascaded} selected $44,820$ images from the Multi-PIE \cite{gross2010multi} as the training set and augmented it with $8,965$ synthetic 2D face images. The authors demonstrated that the face detection error of the cascaded regression (CR) based method significantly decreased, when trained with augmented data (plot in red) compared to when the landmark detection model was trained on only real faces (plot in blue). Motivated by this observation, the augmented data was used to train the proposed method based on cascaded collaborative regression (CCR, plot in black) to achieve the best face detection performance.}
		\label{augmentation_result}
\end{figure} 

   \subsection{Increasing the size and diversity of training dataset} 
   
   Training deep neural networks requires a tremendous amount of data. At the same time,  datasets in human analysis applications have often very limited samples. However, training with smaller datasets may lead to poor generalization of the real-world test examples. Therefore, several studies in human analysis advocate augmentation through synthetic data. Augmentation with synthetic data improves diversity by introducing more variations in training data, as well as increases the size of the training set. Training with a more extensive and diverse set leads to improved training and generalizability of the trained model on the test data.
   
   \par Feng \textit{et al.} \cite{feng2015cascaded} discussed the limited availability of annotated datasets to train a facial landmark detection model. The authors generated $8,965$ synthetic 2D face images to address this limitation with 11 different yaw rotations and five pitch rotations (see Figure \ref{augmentation_synthetic}). The authors augmented the training set for landmark detection and found that the face detection error reduces significantly after training on the augmented dataset (see Figure \ref{augmentation_result}). Similarly, Masi \textit{et al.}  \cite{masi2019face} augmented the training set of face images using augmentations that introduce variations in pose and shape. The authors demonstrated that rank-1 face recognition accuracy on the IJB-A dataset \cite{klare2015pushing} improved from $94.6\%$ to $96.2\%$ after augmentation with synthetic samples. Several other studies additionally advocated augmenting the training set with synthetic data. Some of these studies include applications in human posture recognition \cite{gouiaa2017learning}, brain-based authentication \cite{piplani2018faking}, signature verification \cite{melo2019deep}, face photo-sketch recognition \cite{galea2017forensic,yu2019improving}, face recognition \cite{marriott20213d}, cross spectral face recognition \cite{anghelone2021explainable} and facial expression analysis \cite{bozorgtabar2019using}.

    \begin{figure}
		\centering \includegraphics[scale=1]{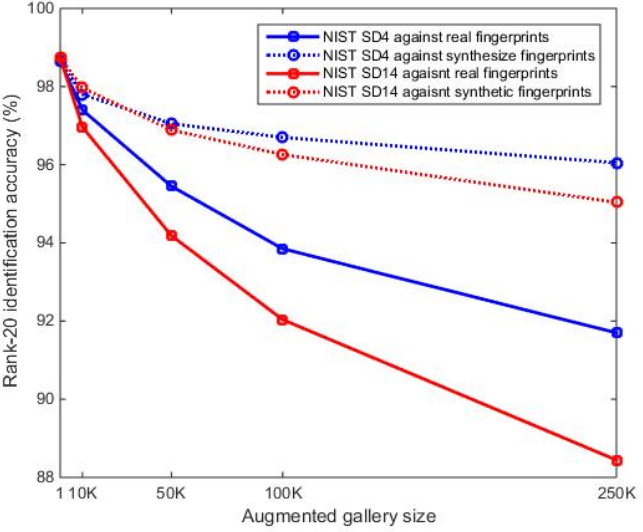}
		\caption{Cao and Jain \cite{cao2018fingerprint} generated synthetic fingerprints and augmented the gallery of standard real fingerprint databases to assess the scalability of state-of-the-art fingerprint search algorithms \cite{cao2017fingerprint}. The authors found that the rank-20 identification accuracies dropped as the gallery was augmented with synthetic fingerprints. These results signified the usefulness of synthetic data, in order to assess the scalability of systems.}
		\label{scalability_fingerprints}
\end{figure} 
   \subsection{Assessing scalability of systems} 
   \label{sec:assess-scalability}
   Evaluation of scalability of large-scale systems such as the Aadhar database maintained by the unique identification authority of India requires assessment of a system's performance for a colossal number of enrollees, sometimes up to a billion (Aadhar has 1.32 billion enrollments till 31 October 2021\footnote{\url{https://uidai.gov.in/}}). Scalability analysis of automated systems is crucial to assess whether these can be deployed for large-scale real-world applications. However, the collection of such large-scale datasets pertaining to humans is often challenging. To address this problem, researchers proposed to generate large-scale synthetic data instead of relying on real large-scale datasets. Such synthetic data is instrumental in performing scalability analyses of human analysis systems. 
   
    We note that the scalability can either be evaluated with system-relevant metrics (\textit{e.g.}, throughput rate) or metrics that reflect the employed algorithms' performance or pre-trained models. According to the work of Sumi \textit{et al.} \cite{sumi2006study}, synthetic evaluation datasets have to comply with following three criteria. 
    \begin{enumerate}
        \item \textbf{Privacy.} There shall not be a link between a synthetic sample to one of the individuals contained in the training dataset.
        
        \item \textbf{Precision.} The performance of a pre-trained model evaluated with synthetic data shall be equal to the performance reported based on real data.
        
        \item \textbf{Universality.} The precision shall be consistent across the evaluation of different pre-trained models. 
    \end{enumerate}
   
   \par Wilson \textit{et al.} \cite{wilson2003studies} demonstrated that the identification performance of a fingerprint recognition system drops linearly with the increase in enrolment records in the gallery. This observation motivated Cao and Jain \cite{cao2018fingerprint} to generate  10 million synthetic rolled fingerprints using I-WGAN \cite{gulrajani2017improved}, in order to evaluate the scalability of fingerprint search algorithms. Similar to the trend observed for real data \cite{wilson2003studies}, the authors found that the rank-20 accuracy on NIST SD4 \footnote{\url{https://www.nist.gov/srd/nist-special-database-4}} accuracy drops from $98.7\%$ to $96.1\%$ after the gallery is augmented with 250K synthetic fingerprints generated by the authors. Related to that, the report NIST SD14 \cite{watson1993nist} indicated that the rank-20 accuracy drops from $98.7\%$ to $95.0\%$ (see Figure \ref{scalability_fingerprints}). 
   
   Recently, Colbois \textit{et al.} \cite{colbois2021use} analysed the verification accuracy and privacy of synthetic face images generated with StyleGAN2 \cite{karras2020analyzing} and InterFaceGAN \cite{shen2020interpreting}. The authors introduced a synthetic version of the Multi-PIE dataset \cite{gross2010multi} (Synth-Multi-PIE), representing the same factors of variation. The precision was assessed following the evaluation protocol of \cite{gross2010multi}, identifying only minor performance differences between Synth-Multi-PIE and Multi-PIE. Similar studies on scalability analysis using synthetic data have been conducted for signature verification \cite{ferrer2014static}, hand-shape biometrics recognition \cite{morales2015synthesis}, face recognition \cite{osadchy2017g}, iris verification \cite{drozdowski2017sic} and keystroke dynamics \cite{migdal2019statistical}.

%   Ferrer \textit{et al.} \cite{ferrer2014static} generate synthetic handwritten signatures to evaluate the performance of a signature verification system for a large number of users. Morales \textit{et al.} \cite{morales2015synthesis} generate synthetic hand-shape images to evaluate scalability of hand-shape biometrics recognition systems.  Osadchy \textit{et al.} \cite{osadchy2017g} evalaute the scalability of a face authentication system (SecureFace \cite{dunkelman2013poster}) through evaluation on large number of synthetic face images. Drozdowski \textit{et al.} propose Sic-Gen \cite{drozdowski2017sic}, synthetic iris template generator to obtain samples that can be used for scalability analysis of large-scale testing of iris verification. Migdal and Rosenberger \cite{migdal2019statistical} propose a statistical model to generate large scale synthetic keystroke dynamics databases.
   
%   Friedman and Komogortsev \cite{friedman2017synthetic} curate synthetic database using eye movement features to assess biometric performance under different variations. 
% Ranjan \textit{et al.} \cite{ranjan2018light} discuss the limitation of state-of-the-art eye gaze tracking systems due to head pose variations. To address this limitation, the authors generate synthetic eye gaze dataset with variations in head pose using a computer graphics tool named SynthesEyes \cite{wood2015rendering}. The model is trained on the synthetic data and then fine-tuned using real data.
   
    \begin{figure}
		\centering \includegraphics[scale=0.15]{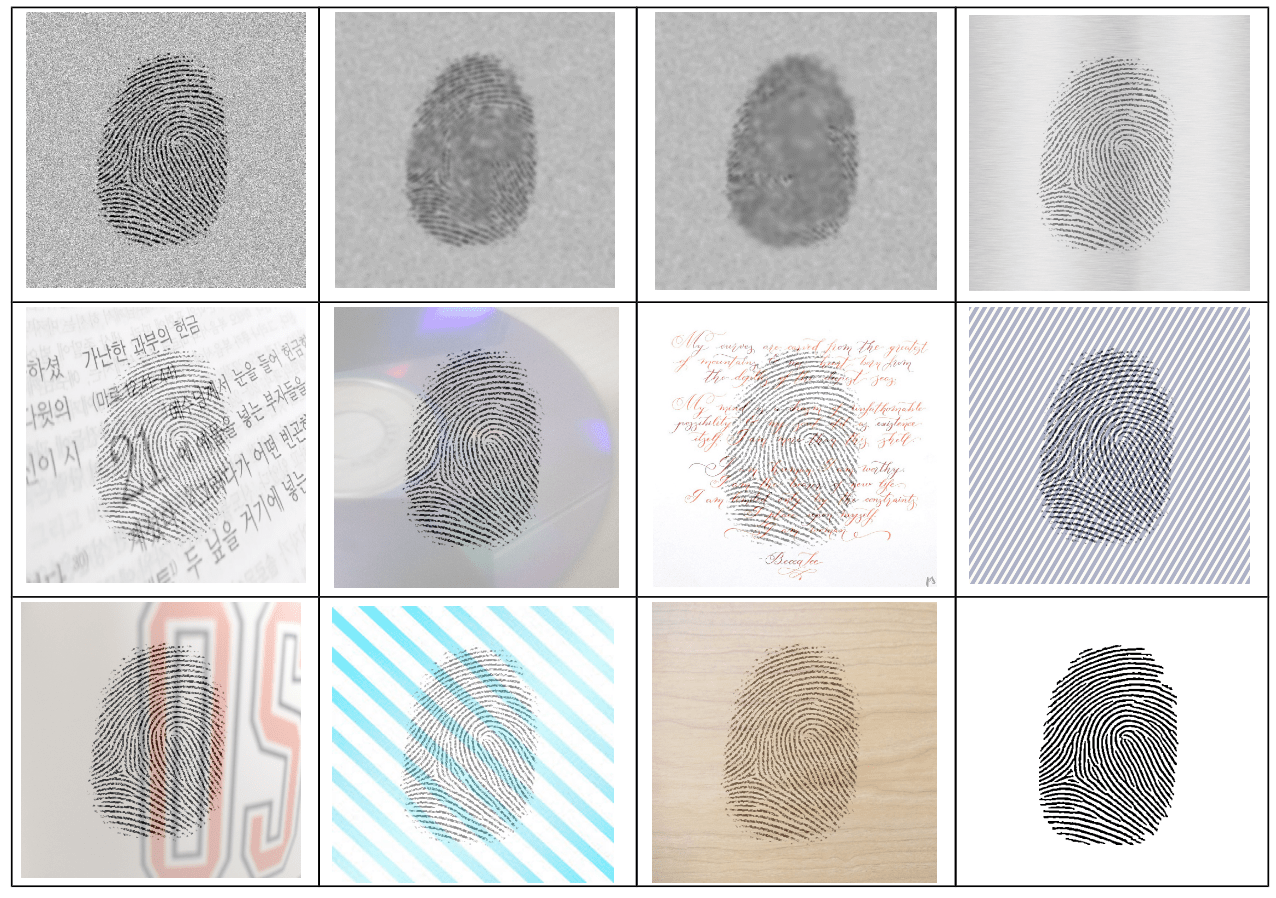}
		\caption{Joshi \textit{et al.} \cite{joshi2019latent} \cite{joshi2021training} argued that a pixel-level annotated training set to train a supervised fingerprint enhancement model was unavailable. To alleviate this challenge, the authors generated a training set of $9,042$ synthetically distorted fingerprints and the corresponding annotations. In this figure, all fingerprints have a common ground truth enhanced image (third row, fourth column). Thus, synthetic data can be used as a proxy dataset for several applications that originally lack an unannotated training dataset, however need it for training the model.}
		\label{synthetic_annotations}
\end{figure}

    \subsection{Providing annotated data for supervision}
    \subsubsection{Supervised Learning}
    Numerous applications can be formulated as a supervised learning problem, however, real annotated data cannot be obtained for them. For such applications, representative synthetic samples and their annotations are generated in order to train models in supervised learning manner \cite{cao2015latent,richardson20163d,varol2017learning,basak2020methodology,park2018learning,roberto2017procedural,  xu2020augmentation,ibsen2022face,anghelone22thermal,richardson2017learning} (see Figure \ref{synthetic_annotations}). Feng \textit{et al.} \cite{feng2015cascaded} argued that manually annotated facial landmarks are often inaccurate for occluded facial regions. While the annotations of synthetic faces generated from a 3D model are correct for all different pose variations as these are direct projections to 2D from 3D. Therefore, the authors used a synthetic dataset to obtain reliable and consistent annotations for various image variations. Similarly, Liu \textit{et al.} \cite{liu20193d} employed synthetic data with dense point-to-point correspondence maps towards learning a 3D face model. Some applications have exploited synthetic data to learn a transformation from distorted to clean samples. Associated to this direction, Dieckmann \textit{et al.} \cite{dieckmann2019fingerprint} proposed to learn the pre-aligning of fingerprint images through horizontally and vertically translated and rotated synthetic fingerprints. Likewise, Zhang \textit{et al.}  \cite{zhang2016multi}, Joshi \textit{et al.} \cite{joshi2019latent} and Nojavanasghari \textit{et al.} \cite{nojavanasghari2017hand2face} utilized synthetic data to learn blind inpainting of face images, enhancement of fingerprints and transformation from occluded to non-occluded faces, respectively.

\begin{figure}
		\centering \includegraphics[scale=0.67]{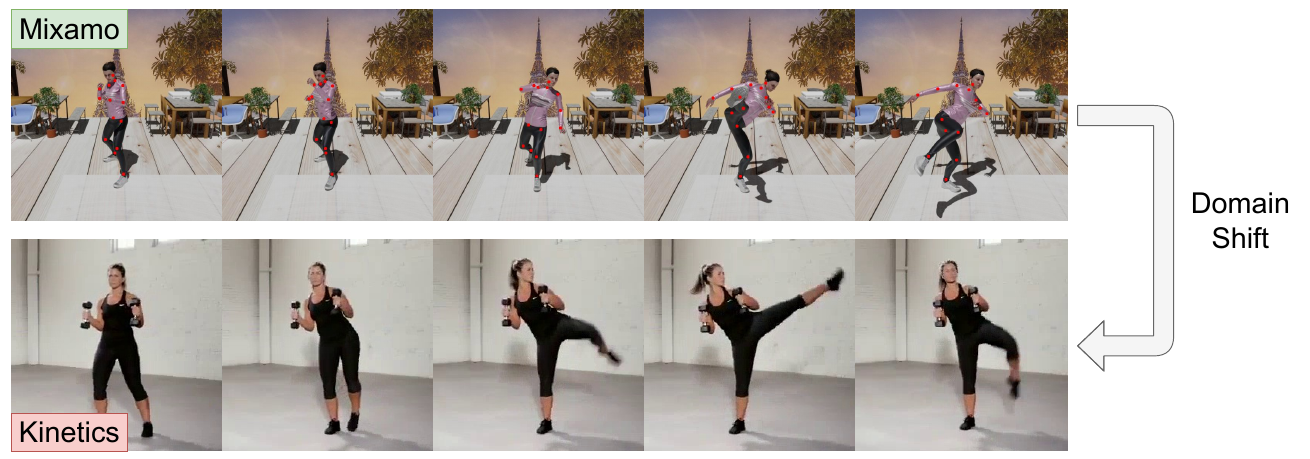}
		\caption{Costa \textit{et al.} \cite{da2022dual} introduced a large-scale synthetic human action recognition dataset to promote the design of unsupervised domain adaptation methods for minimizing the cost and human effort in acquiring a large annotated dataset for human action recognition.}
		\label{synthetic_domain daptation}
\end{figure}

% \begin{figure}
% \centering
% \includegraphics[scale=0.5]{images/domain_adaptation_synthetic.pdf}
% \caption{Joshi \textit{et al.} \cite{joshi2021sensor} highlight the dependence of state-of-the-art fingerprint segmentation models on annotated data to obtain satisfactory performance on a newly introduced fingerprint sensor. To eliminate this need, the authors propose to exploit annotated synthetic data to train the model. To close the performance gap on the new sensor, the authors propose feature alignment based model adaptation using only the unannotated samples from the new sensor. Thus, training on synthetic data and domain adaptation on the unannotated real data helps to eliminate the need for annotations of the real data.}
% \label{domain_adaptation_synthetic}
% \end{figure}
    
    \subsubsection{Unsupervised domain adaptation}
    \label{sec:unsupervised-domain-adaption}
    Supervised deep neural networks require a massive amount of manually annotated training data. However, collection, and particularly annotation of such is tedious, time-consuming and expensive. Furthermore, many human analysis applications require annotations by domain experts \cite{joshi2021sensor}, or reliable annotations cannot be obtained for the real data \cite{bondi2020birdsai}. To address this challenge, researchers proposed to train models on a synthetic training dataset whose annotations can be computationally acquired. However, a huge gap in model performance was observed between real and synthetic data due to the visible \textit{domain shift} (see Figure \ref{synthetic_domain daptation}). Researchers adapted models to unannotated real-world datasets, in order to reduce the performance gap between real and synthetic data. An important application of unsupervised domain adaptation of human analysis models includes the contributions of Wang \textit{et al.} \cite{wang2021pixel} \cite{wang2019learning}. The authors exploited $15,212$  synthetic labelled crowd scene images containing more than $7,000,000$ subjects for the purpose of training a model for pixel-level understanding in a crowd. However, instead of directly using the synthetic data, the authors firstly translated synthetic images into realistic images using a GAN. This was beneficial in reducing the domain gap between synthetic and real data. Next, the model was trained on translated images instead of actual real images. The authors reported that the structural similarity index measure (SSIM) value improved from $0.554$ to $0.660$ after exploiting the synthetic crowd counting dataset. 
    
    \par Joshi \textit{et al.} \cite{joshi2021sensor} highlighted the dependence of state-of-the-art fingerprint segmentation models on annotated data as a means to obtain satisfactory performance on a newly introduced fingerprint capture device. To mitigate this limitation, the authors only used  synthetic data (source domain) annotations to learn fingerprint segmentation. To adapt the model to a new fingerprint capture device (target domain), the authors aligned the source and target domain features using recurrent adversarial learning. Extending the theme of unsupervised domain adaptation, Bondi \textit{et al.} \cite{bondi2020birdsai} argued that annotations of thermal infrared videos were often erroneous and therefore proposed to train the detection and tracking model on a synthetic dataset, adapting subsequently to a real dataset. Several applications spanning areas such as face recognition \cite{zhong2020depth}, person re-identification \cite{bak2018domain}, human action recognition \cite{da2022dual} and head pose estimation \cite{kuhnke2019deep} successfully exploited synthetic data to eliminate the need for annotations of real data through unsupervised domain adaptation.
   
%   Zhong \textit{et al.} \cite{zhong2020depth}
%     \textcolor{red}{Read new reference}\cite{bak2018domain} Cozzolino \textit{et al.} \cite{cozzolino2021id}  

    % \subsection{Improves generalization ability (through Pre-training)} 
    
    \begin{figure}
		\centering \includegraphics[scale=0.85]{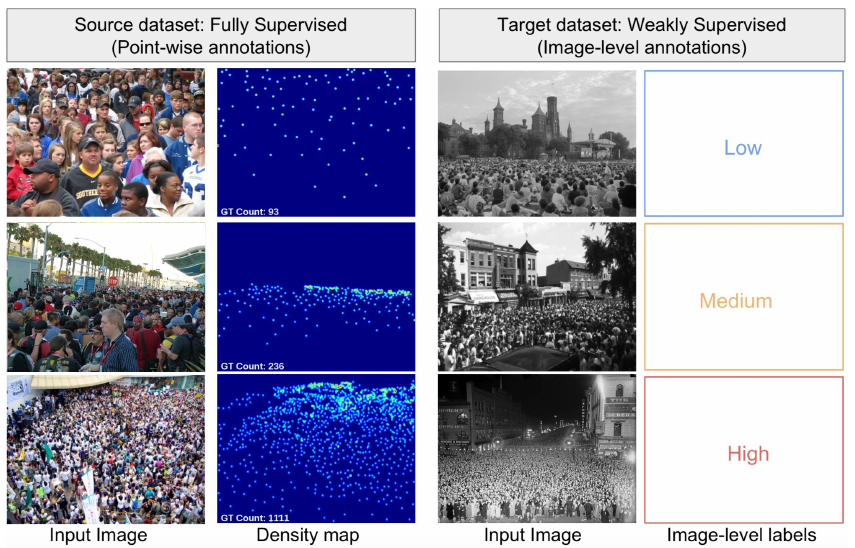}
		\caption{Sindagi \textit{et al.} \cite{sindagi2019ha} studied domain adaptation of a crowd counting model. While the source had pixel-level annotations, the target data was annotated on image level and only provided weak supervision.}
		\label{weak_supervision_example}
\end{figure}

 \begin{figure}
		\centering \includegraphics[scale=0.55]{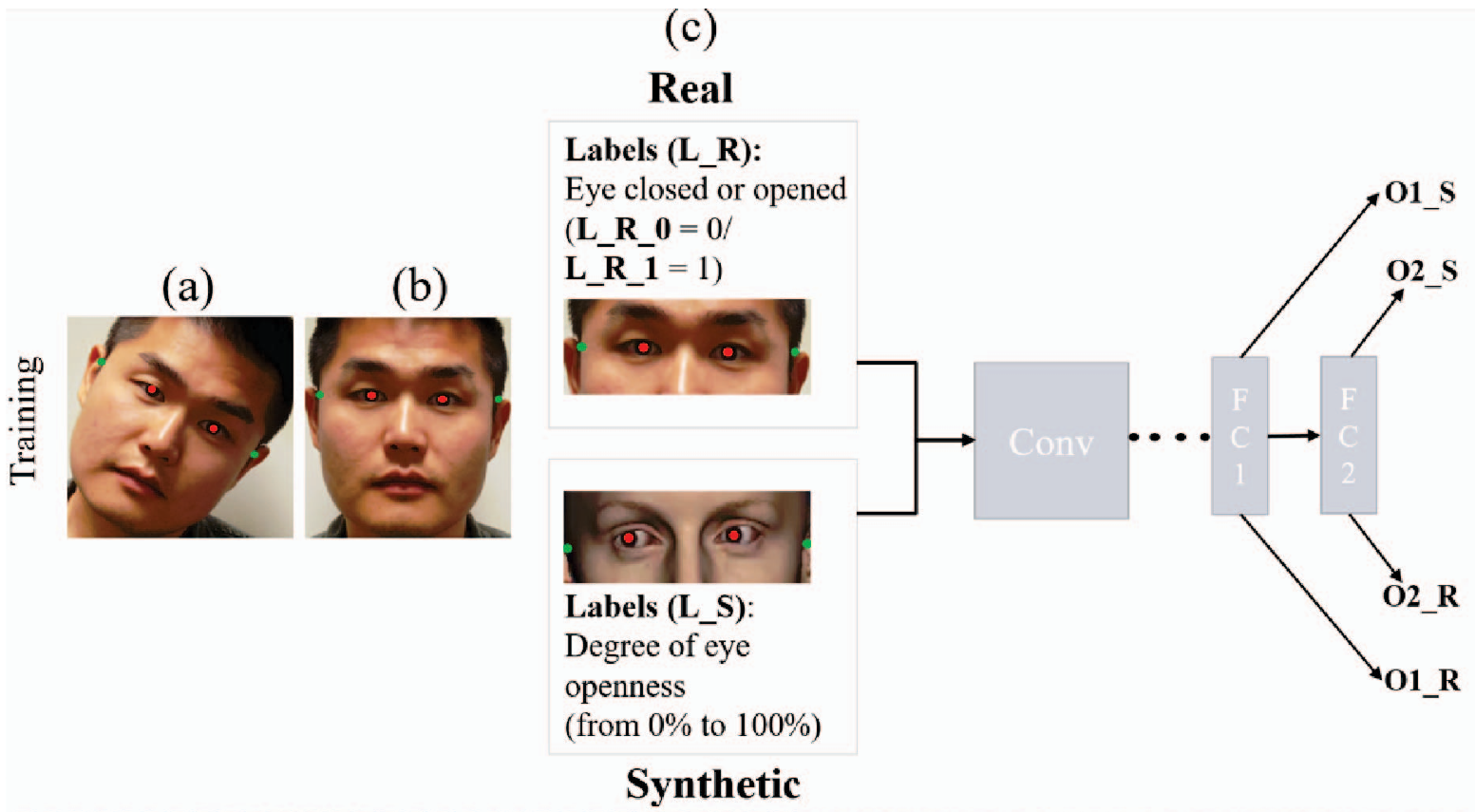}
		\caption{Mequanint \textit{et al.} \cite{mequanint2019weakly} proposed weakly supervised learning in an eye-closeness estimation model. The model exploited synthetic annotated data that provided a degree of openness of eyes, whereas the real data only provided weak supervision, whether the eye is open or closed. Thus, annotated synthetic data can be used to enable learning in weakly supervised learning.}
		\label{augmentation_weak_supervision}
\end{figure}
    
    \subsubsection{Weakly supervised learning}
    \label{sec:weak-supervised-learning}
    %These approaches are designed for scenarios when only weak supervision is available for training the human analysis model (see Figure \ref{weak_supervision_example}). 
    Synthetic annotated data has been utilized in weakly supervised learning (see Figure \ref{weak_supervision_example}) aiming to introduce a higher degree of supervision. %, which is impossible with weakly supervised annotations. 
    For instance, Mequanint \textit{et al.} \cite{mequanint2019weakly} highlighted the unavailability of annotated data for training an eye-openness estimation model. To alleviate this issue, the authors generated $1.3$ million annotated synthetic eye images with varying levels of eye openness to enable supervised learning. Furthermore, to counter the domain shift between real and synthetic eye images, the authors exploited weak supervision (eyes simply open or closed). It was demonstrated that the classification (open/close) accuracy improves from $96.30\%$ to 100\% after utilizing synthetic data. Deviating from the above, Zhang \textit{et al.} \cite{zhang2019facial} generated weakly labelled face images (labels as bounding box and class) using a deep convolutional generative adversarial network (DCGAN) \cite{radford2015unsupervised} and used a limited amount of fully annotated real data (labels as landmark vector, bounding box and class). A weakly supervised learning framework was used to train the facial landmark detection model, which improved the average error distance for landmark detection on the labelled face parts in the wild (LFPW) dataset \cite{belhumeur2013localizing} from $4.25$ to $3.12$ after utilizing synthetic faces.

    % not using synthetic data \cite{huang2020eye} \cite{chen2021weakly} \cite{peng2019weakly} 
    
    \subsection{Pre-training a deep model}
    \label{sec:pretrain-deep-model}
    Deep neural network models impart a large number of parameters and, therefore, require a large amount of training data to avoid over-fitting. We have that the ImageNet dataset incorporates approximately $1.2$ million annotated images. However, in human analysis often only limited annotated training sets are publicly available, including \textit{e.g.,} hundreds or thousands of images. Therefore, once again synthetic data is advantageous in alleviating the need for a large amount of training data required for training data-hungry deep models. It is common practice to generate annotated synthetic datasets and use such to pre-train deep models, which are then fine-tuned with annotated real data. A number of studies demonstrated that such pre-training with synthetic datasets leads to better performance than training directly on the real dataset. In one of the recent studies, Engelsma \textit{et al.} \cite{engelsma2022printsgan} demonstrated that performance gain was observed by a DNN-based fingerprint recognition model (DeepPrint) \cite{engelsma2019learning} that was pre-trained on synthetic fingerprints and fine-tuned on real fingerprints. The authors generated $525$K synthetic fingerprints for pre-training DeepPrint and fine-tuned it on $25$K fingerprints from the NIST SD302 database \cite{fiumara2019nist}. The authors then assessed the fingerprint recognition performance of DeepPrint on NIST SD4 database\footnote{\url{https://www.nist.gov/srd/nist-special-database-4}}, with and without pre-training with synthetic data. The authors observed that the true acceptance rate (TAR) @ false acceptance rate (FAR)=$0.01\%$ improves from $73.37\%$ to $87.03\%$, when pre-trained with synthetically generated fingerprints. 
    
    \par Similarly, Wang \textit{et al.} \cite{wang2021pixel} trained a pixel-level crowd understanding model on large-scale synthetic data ($15,212$ images of $7,625,843$ subjects) and fine-tuned it on labelled real data. The mean square error decreased by $14.1\%$ after pre-training on synthetic data was noted, compared to the performance of the crowd counting model pre-trained on ImageNet dataset \cite{deng2009large}. Similar trends %as reported in \cite{wang2021pixel} and \cite{engelsma2022printsgan} 
    were observed for other applications analyzing human data including speech recognition \cite{bird2020lstm}, hand shape recognition \cite{svoboda2020clustered}, head pose estimation \cite{basak2021learning}, eye gaze tracking \cite{ranjan2018light}, re-identification of individuals \cite{barbosa2018looking} and face recognition \cite{kortylewski2018training,kazemi2018attribute}.

    \subsection{Fine-tuning a pre-trained model}
    \label{sec:fine-tune-pre-trained-model}
    In addition to pre-training with synthetic data, the following counteracts the need for large-scale annotated data. % for training a deep model, researchers often initialize 
    The parameters of deep neural networks are initialized with pre-trained weights of standard deep models trained on large-scale datasets of real images. Such networks are then fine-tuned with annotated synthetic dataset associated to the respective application. Following this approach, Dou \textit{et al.} \cite{dou2017end} proposed a 3D face reconstruction model, whose weights were initialized with the parameters of the VGG-Face model \cite{Parkhi15} (a standard model trained on real human faces) and which was then fine-tuned on $250$K synthetic $2D$ face images. Kim \textit{et al.} \cite{kim2020dcnn} pre-trained the near-infrared (NIR) face recognition model on $453,414$ RGB face images of the CASIA WebFace dataset \cite{yi2014learning}. Next, $32,992$ synthetic NIR face images were generated using CycleGAN \cite{zhu2017unpaired}, augmenting the training set for fine-tuning. The authors reported a $0.31\%$ increase in face identification rate after fine-tuning the augmented data compared to training the model from scratch. Similar applications of synthetic data for fine-tuning of a pre-trained model include finger vein recognition \cite{ou2021fusion}, iris PAD \cite{fang2022overlapping} and fingerprint PAD \cite{rohrer2021gan}.
    
    % \cite{wang2018generative}

    % \item \textbf{Generate a balanced dataset:}

    % \par An \textit{et al.} \cite{an2017multi}. The authors
    % propose a framework for person re-identification that generates synthetic samples on intermediate levels to account for the feature gap in different views and achieve better generalizability.  The authors demonstrate that the task of synthetic data generation improves representation ability of the model. Deb and Guirguis \cite{deb2020use} learn the distribution of touchstrokes using AC-GAN by generating synthetic samples. Subsequently, the authors demonstrate the ability of the trained discriminator as a touchstroke authentication model that can differentiate between real and synthetic touchstrokes.  

    % \subsection{Mitigates lack of availability of data}

   \subsection{Enforcing consistency regularization}
   
   \subsubsection{Contrastive Learning}
   Contrastive learning is a learning paradigm that ensures that representations of similar samples must be close, whereas representations of dissimilar samples are far apart in the latent space. Various studies exploited synthetically augmented data to generate similar samples for a given input. Subsequently, using contrastive learning, the model was encouraged to have similar representations for the original and the augmented input samples. %   Ruiz \textit{et al.} \cite{ruiz2020off} exploit transformations such as morphological dilations, rotation and addition to noise (see Figure \ref{consistency_regularization_contrastive}) as well mathematical modelling to generate synthetic genuine pairs of signatures. Subsequently, the authors the synthetic signatures for training a Siamese network for offline handwritten-signature verification. The authors demonstrate that the equal error rate (EER) improves from 11.11\% to 4.90\% after exploiting synthetic data.
   Ryoo \textit{et al.} \cite{ryoo2018extreme} introduced different low resolution (LR) transformations into videos and trained an activity recognition model such that the images obtained from the same scene, pertaining to different pixel values due to LR transformation shared a common embedding. The authors demonstrated that the classification accuracy under low-resolution constraints improves from $31.50\%$ to $37.70\%$ after using synthetic data. Neto \textit{et al.} \cite{neto2021focusface} applied augmentation techniques to generate synthetically masked faces. Contrastive learning brought then representations of masked and unmasked faces of the same data subject close to each other. The authors demonstrated that the model trained using the synthetically masked images outperformed existing standard face recognition systems on masked face recognition. Several other applications in speaker recognition \cite{huh2020augmentation}, face recognition \cite{lee2020learning}, person re-identification \cite{chen2021joint} and  electrocardiogram (ECG) based authentication \cite{pinto2020self} proposed to generate synthetic data for exploiting contrastive learning.

\begin{figure}
		\centering \includegraphics[scale=0.9]{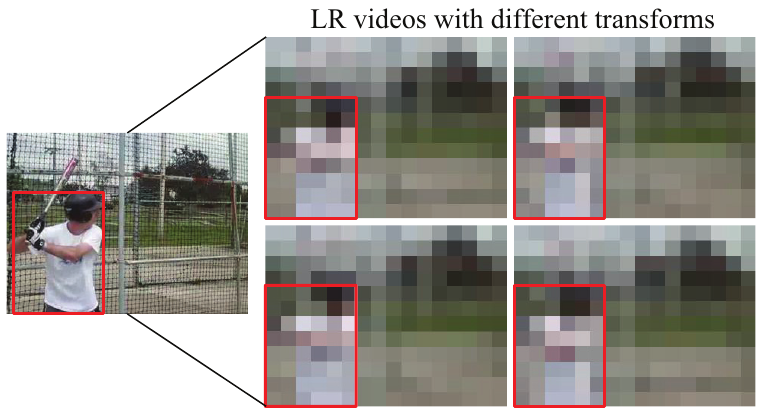}
		\caption{Ryoo \textit{et al.} \cite{ryoo2018extreme} exploited different low-resolution transformations towards synthetically generating videos with the same scene and different pixel values due to changes in resolution. Later, a Siamese network was trained that ensures that the feature representations of the original and augmented video frames were similar. Thus, synthetic data can be used to ensure consistency among representations learnt by a human analysis model.}
		\label{consistency_regularization_contrastive_action_recognition}
\end{figure}

\begin{figure}
		\centering \includegraphics[scale=1.12]{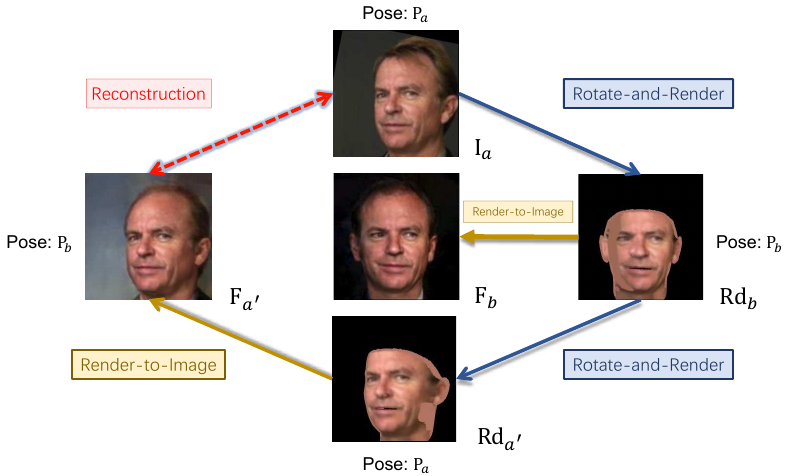}
		\caption{Zhou \textit{et al.} \cite{zhou2020rotate} proposed rotate-and-render augmentation that given a face image generates a synthetic face image with a varying pose. Later, the synthetic face was rendered back to the original pose. Such an augmentation ensured self-supervision in face recognition models. As a result, the model learned to preserve consistency in identity information while varying facial poses.}
		\label{augmentation_self-supervision}
\end{figure}

   \subsubsection{Self-supervision} 
   Self-supervision is an unsupervised learning paradigm 
%   in which the learning problem is posed in a supervised manner by introducing some auxiliary task for which (pseudo) labels can be generated.
   through which a model can be regularized by introducing an auxiliary task. Several approaches in human analysis have introduced transformations to an input data to generate synthetic labelled data for training the auxiliary task in a supervised manner. For example Zhou \textit{et al.} \cite{zhou2020rotate} proposed rotate-and-render, an augmentation technique that rotates faces back and forth in $3D$ space and subsequently renders them back in $2D$ (see Figure \ref{augmentation_self-supervision}). %The authors demonstrate that the rotate-and-render 
   Such augmentation strategy ensured consistency regularization, while training face recognition models. As a result, $TAR@FAR=0.001$ on the IJB-A dataset improved from $80.00\%$ to $82.48\%$ after introducing self-supervision through the proposed augmentation strategy. Other applications utilizing synthetic data for self-supervision include deepfake detection \cite{zhao2021learning}, facial expression recognition \cite{li2021self}, face recognition \cite{ju2022complete} and sleep recognition \cite{zhao2020self}.

%   Liu \textit{et al.} \cite{liu2021taming} propose to improve generalization ability of a presentation attack detector for unseen presentation attack instruments using self-supervised learning. The authors propose to exploit De-Mixing and De-Folding as the pretext tasks. Detector trained with these pretext tasks obtain better generalization on unseen attacks.
 
   \begin{figure}
		\centering \includegraphics[scale=0.7]{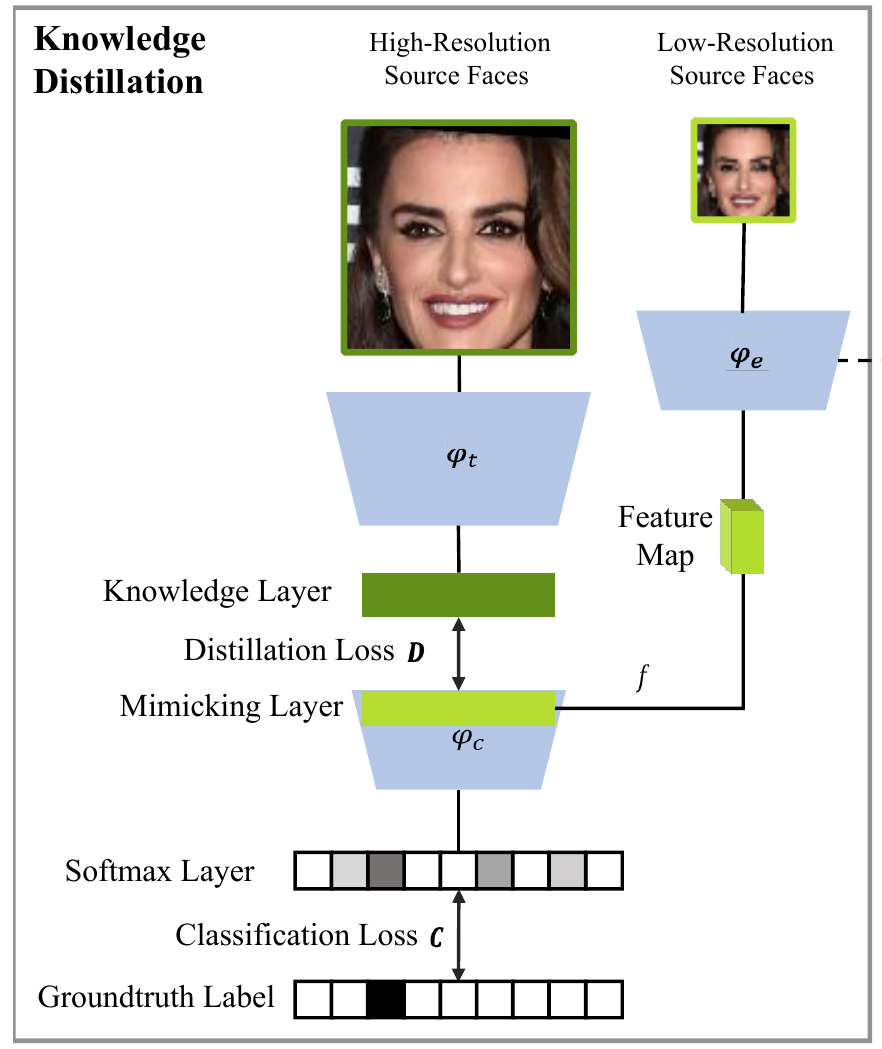}
		\caption{Ge \textit{et al.} \cite{ge2019fewer} proposed a knowledge distillation framework for few-shot face recognition in the wild. The authors exploited consistency regularization among the output of the teacher model for the high-quality input and output of the student model for synthetic low-quality face images. Therefore, synthetic data can be used to enforce consistency regularization for improved performance of the human analysis model in few-shot learning scenarios.}
		\label{augmentation_few_shot}
\end{figure}

   \subsubsection{Few-shot learning} 
   
   Few-shot learning is characterized by learning with a limited number of samples. Specifically, in order to compensate for limited availability of data and promote the \textit{learning to learn} paradigm, %few-shot learning-based studies in human analysis use 
   augmentation strategies simulate challenging real-world scenarios and ensure consistency in prediction for real and augmented input sample. Ge \textit{et al.} \cite{ge2019fewer} proposed in this context a knowledge distillation framework to improve face recognition performance under limited data and low resolution constraints. The face recognition model was trained on high-resolution face images, serving as teacher network. The authors then synthetically generated low-resolution face images and trained the student model such that the output of the student model on the synthetic low-resolution face was close to the output of the teacher model on the real high-quality face image (see Figure \ref{augmentation_few_shot}). The associated performance of face verification on the UMDFace dataset \cite{bansal2017umdfaces} improved from $67.59\%$ to $73.58\%$ after knowledge distillation compared to training the student model directly on synthetic faces. Thus, consistency regularization between real and synthetic data improved the face recognition performance with few-shot learning. Other studies utilizing synthetic data for few-shot learning in human analysis include applications in attribute-based person search \cite{cao2020symbiotic}, deepfake detection \cite{korshunov2022improving}, login authentication \cite{solano2020few}, signature verification \cite{tolosana2021deepwritesyn}, speaker recognition \cite{li2020automatic}  and gaze estimation \cite{yu2019improving}.

\begin{figure*}
\centering
\includegraphics[width=\textwidth]{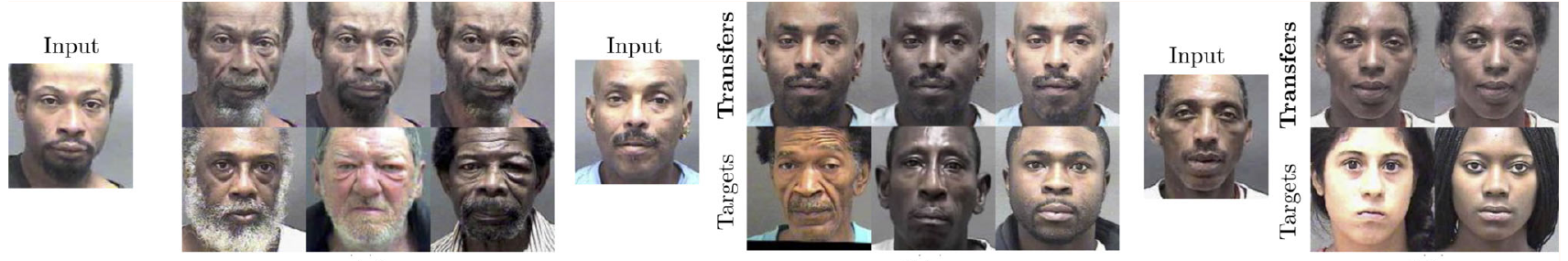}
\caption{Georgopoulos \textit{et al.} \cite{georgopoulos2021mitigating} improved intra-class diversity in the training set by transferring demographic attributes (left to right): age, skin tone and gender. The authors demonstrated that training with diverse synthetic samples of the same subject is instrumental in mitigating demographic bias observed in face recognition models.}
\label{bias_synthetic1}
\end{figure*}  

\begin{figure*}
\centering
\includegraphics[width=\textwidth]{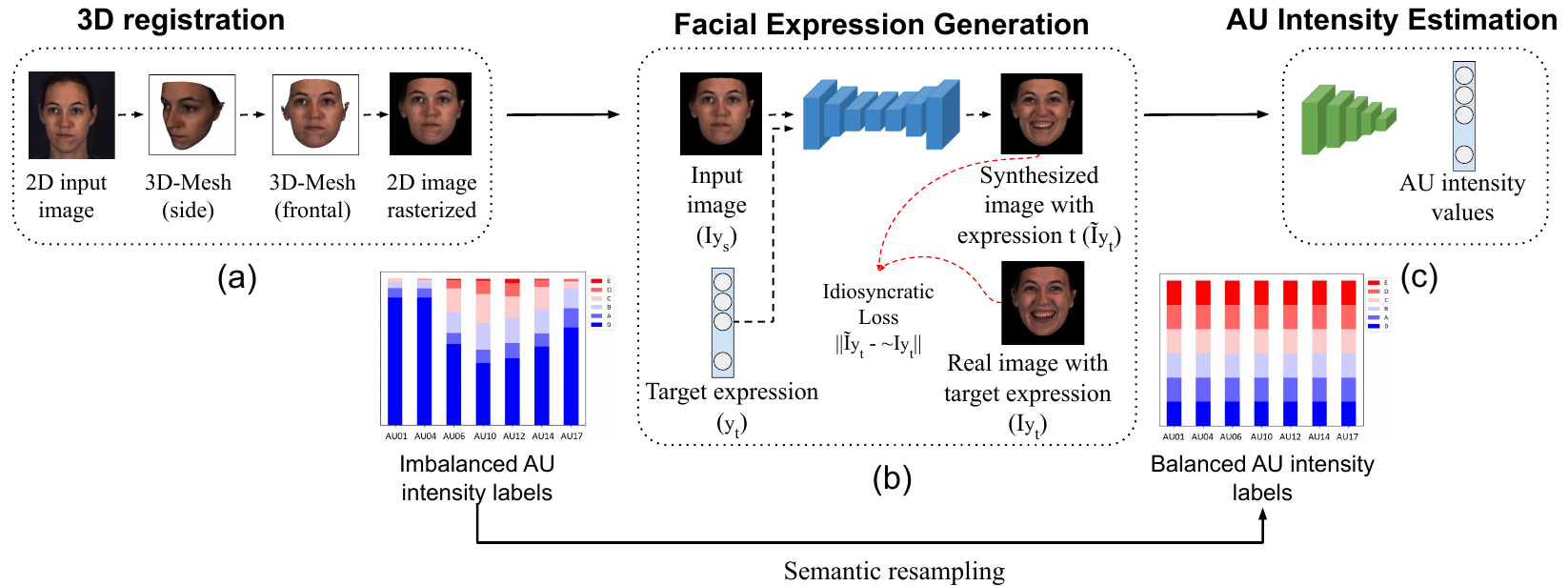}
\caption{Niinuma \textit{et al.} \cite{niinuma2021synthetic} generated synthetic faces, in order to create a training set that is balanced \textit{w.r.t.} action unit (AU) intensity labels. The authors demonstrated that training on the balanced training set improved detection of facial actions.}
\label{bias_synthetic2}
\end{figure*}  
    \subsection{Mitigating dataset bias and ensuring fairness} %Training a deep neural network for human analysis often requires a large training database. However, the collected training 
    Human datasets often contain demographic bias \textit{w.r.t.} attributes such as ethnicity, gender, or age \cite{drozdowski2020demographic}. In addition, collected datasets might be biased to a certain group of labels \cite{niinuma2021synthetic}. Synthetic data is able to balance and unbias datasets beneficial in training and designing fair and unbiased human analysis models. Georgopoulos \textit{et al.} \cite{georgopoulos2021mitigating} exploited an attribute-transfer based approach to balance underrepresented demographic groups in training datasets. Attributes such as skin tone, gender, and age were transferred into given training samples (see Figure \ref{bias_synthetic1}) towards creation of an unbiased training dataset. In the related study the accuracy of face recognition on dark-skinned women over $60$ years old characterized by true positive rate (TPR) improved by $20\%$ on the UNCW dataset \cite{ricanek2006morph} after training on the training set augmented with synthetic faces.
    
    \par Similarly, Niinuma \textit{et al.} \cite{niinuma2021synthetic} discussed that real datasets employed for facial action detection are not balanced \textit{w.r.t.} action unit (AU) intensity labels. To address this limitation, the authors generates a balanced training set using GANimation \cite{pumarola2020ganimation} (see Figure \ref{bias_synthetic2}). The generated balanced training dataset was used to train the facial action detector, with the related inter-rater reliability score of AU intensity level estimation improving from $48.90\%$ to $52.50\%$ after training the model on synthetic data, as opposed to training on real data. Several other studies in face recognition \cite{kortylewski2019analyzing} \cite{kortylewski2018empirically} \cite{de2020fairness} \cite{zhai2021demodalizing}
    \cite{mcduff2021synthetic} confirmed the ability of synthetic data to train unbiased and fair models. 
    
    %  \cite{de2020fairness} \cite{georgopoulos2021mitigating}
    % \cite{mcduff2021synthetic}. Kortylewski \textit{et al.} \cite{kortylewski2019analyzing} propose augmenting the training set with synthetic samples with different facial poses and identities to have enough representative samples and mitigate dataset bias observed in facial recognition.  McDuff \textit{et al.} \cite{mcduff2021synthetic} generate synthetic data of facial avatars with a variety of facial motions, appearance, background and illuminations conditions. The authors show that training with synthetic dark skin avatars achieves higher generalization ability as compared to the light skin avatars. Pereira \textit{et al.} \cite{de2020fairness} use synthetic data to generate samples that facilate simulation of fair and unfair biometric systems with respect to race and gender. Subsequently, they propose a metric to quantify fairness of a face recognition system.
    
    % \subsubsection{To balance a training dataset}  Zhai \textit{et al.} \cite{zhai2021demodalizing} use Cycle-GAN and PRNet to generate synthetic faces with different poses for improving performance of face recognition model.

\begin{figure}
		\centering \includegraphics[scale=1]{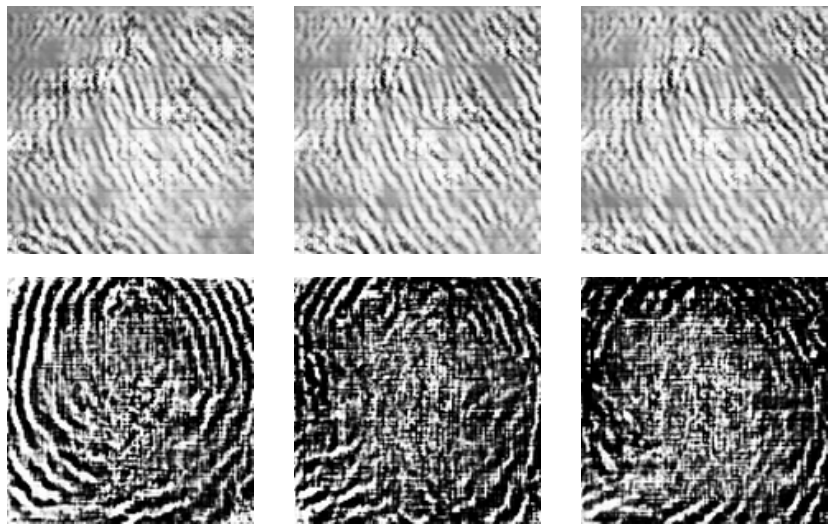}
		\caption{DeepMasterPrints \cite{bontrager2018deepmasterprints} by Bontrager \textit{et al.} constitutes a synthetic fingerprint \textit{masterprint} aimed at presentation attacks on fingerprint recognition systems. The top and bottom illustrate the masterprints for the rolled fingerprints and fingerprints acquired using a capacitive capture device. The first, second and third columns represent the masterprint to achieve a false match rate (FMR) of $0.01\%$, $0.1\%$, and $1\%$, respectively.}
		\label{presentation_attack_example}
\end{figure}

\subsection{Inducing digital perturbation attacks}

   Synthetic data is particularly instrumental in creating novel attacks on biometric systems.
%   \cite{kohli2017synthetic}  \cite{bontrager2018deepmasterprints} \cite{damer2019realistic} \cite{zhang2021mipgan} \cite{damer2021regenmorph} \cite{yadav2019synthesizing} . 
One prominent study in this context constitutes DeepMasterPrints by Bontrager \textit{et al.} \cite{bontrager2018deepmasterprints}, which aimed to generate one masterprint, namely a synthetic fingerprint that was designed to impersonate a set of fingerprints and falsely match with a large number of non-mated enrollees in the enrolment database (see Figure \ref{presentation_attack_example}). This presentation attack for fingerprint recognition systems employed GAN, where the latent input variables in the generator network were obtained using a covariance matrix adaptation evolution technique. The associated false match rate (FMR) of $0.1\%$ increased via DeepMasterPrints to $8.61\%$ on the NIST 9 dataset \cite{watson1993nist}, as well as to $22.50\%$ on the FingerPass DB7 dataset \cite{jia2012cross}. Additional attacks facilitated by synthetic data include those in iris recognition \cite{yadav2019synthesizing,kohli2017synthetic,boutros2020iris}, face recognition \cite{damer2019realistic, nguyen2020generating} \cite{zhang2021mipgan} \cite{damer2021regenmorph} and  fingerprint recognition \cite{bouzaglo2022synthesis}.
   
%   Kohli \textit{et al.} propose iDCGAN \cite{kohli2017synthetic} to generate relistic synthetic iris samples for presentation attack on a commercial iris recognition tool VeriEye \cite{verieye}.   Damer \textit{et al.} propose EMorGAN \cite{damer2019realistic} that is principled on a cascaded enhancement framework to generate high quality images for face morphing attack. On the other hand, MIPGAN \cite{zhang2021mipgan} is a StyleGAN based model that utilizes perceptual and identity losses to generate morphed faces. Different from EMorGAN and MIPGAN, ReGenMorph \cite{damer2021regenmorph} first uses an existing landmark based attack approach to generate morphed images and then trains a GAN to regenerate them as realistic face images that are used for the attack. Yadav \textit{et al.} \cite{yadav2019synthesizing} propose RaSGAN to generate synthetic iris samples and demonstrate the ability of the synthetic images to attack an iris verification system. Bouzaglo and  Keller \cite{bouzaglo2022synthesis} generate 100k pair of synthetic fingerprints and utilize the synthetic fingerprints to spoof fingerprint verification system.

\begin{figure*}
\centering
\includegraphics[width=\textwidth]{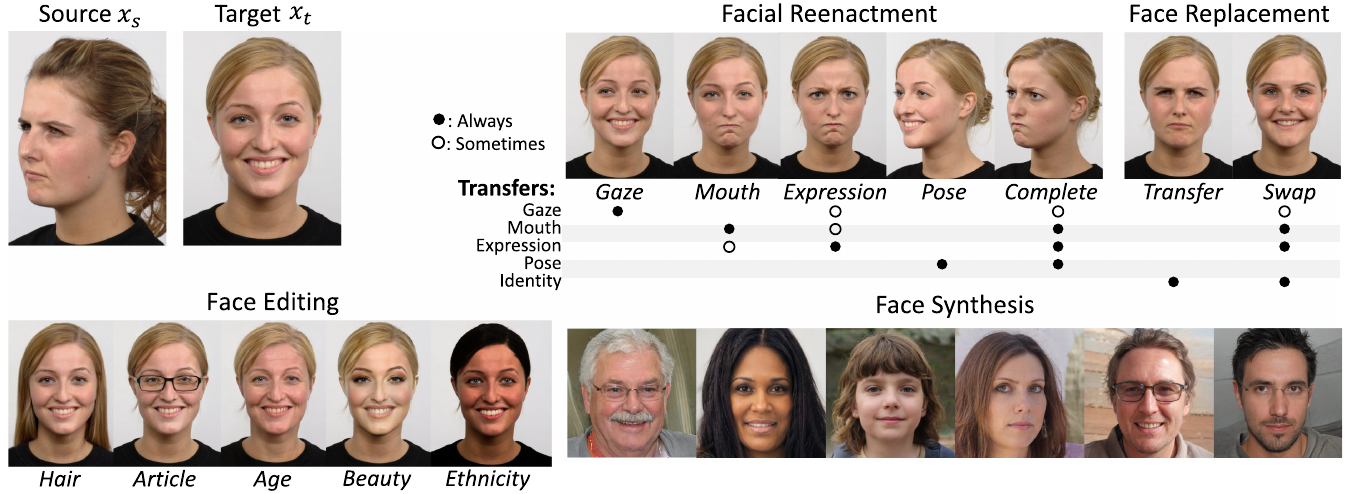}
\caption{Synthetic videos called Deepfakes with varying attributes, can be generated with th induce an attack to ruin the public perception of an individual \cite{mirsky2021creation}.}
\label{deep_fake_example}
\end{figure*}  
    
A related direction has to do with digital human creation \cite{karras2021alias,karras2019style,Karras2019stylegan2,saito2017temporal,tulyakov2017mocogan,wang2020g3an,wang2021inmodegan}, as well as with manipulation of human faces \cite{thies2019deferred,siarohin2019first,wang2022latent}. Specifically, 
a face image of a target individual being superimposed on a video of a source individual has been widely accepted and referred to as \textit{deepfake} (see Figure \ref{deep_fake_example}). Deepfakes entail several challenges and threats, given that (a) such manipulations can fabricate
animations of subjects involved in actions that have not taken place and (b) such
manipulated data can be circumvented nowadays rapidly via social media. Deepfakes are considered in human analysis as digital perturbation attacks, attracting large interest by their own right, with overview articles focusing on deepfake creation and detection \cite{mirsky2021creation,tolosana2020deepfakes,rathgeb2021handbook}, as well as adversarial attacks and defences in images, graphs, and text \cite{xu2020adversarial}. We note that similarly morphing attacks can be introduced using synthetic data \cite{raja2020morphing}. A morphing attack is characterized by a synthetic image for which the authentication system is compelled to match with two contributing subjects instead of one. A morphed image is usually generated by aligning and blending images of two different contributors. For a comprehensive survey on published morphing attacks and associated detection methods, we refer to
related overview articles \cite{scherhag2017biometric,venkatesh2021face,scherhag2019face}.

   %   Therefore deepfakes coerce the target person in a video to reenact the dynamics of the source person.
   
%   is referred to all digital fake content created by means of deep learning techniques

  %  [1, 12]. It was originated after a Reddit user named “deepfakes” claimed in late 2017 to have developed a machine learning algorithm that helped him to swap celebrity faces into porn videos [13]. The most harmful usages of DeepFakes include fake pornography, fake news, hoaxes, and financial fraud. 
  
%   \textcolor{red}{Can we include deep fakes?, morphing attacks?} 

% \subsection{To improve robustness to attacks:} 
% % Li \textit{et al.} \cite{li2020automatic} argue that automated speaker recognition systems often contain limited training data for fresh users. To counter the limited availability of training data, the authors propose a few-shot learning based model to learn representations from limited data. To further improve the generalization ability of the model, the 
% % authors generate synthetic samples through adversarial perturbations. Training the model in this way improves its generalization and robustness compared to state-of-the-art speaker recognition models. 
% \cite{mingxing2021towards} \cite{wu2021improving} \cite{liu2021boosting} \cite{zhang2021empirical} \cite{cai2017countermeasures} \cite{wu2021toward}
%  \cite{chen2021ur} \cite{marriott2020data} \cite{monteiro2020multi} 
% %\end{enumerate}
 
 %not related to synthetic data \cite{seibold2020accurate}
 
 \begin{figure*}
\centering
\includegraphics[scale=0.85]{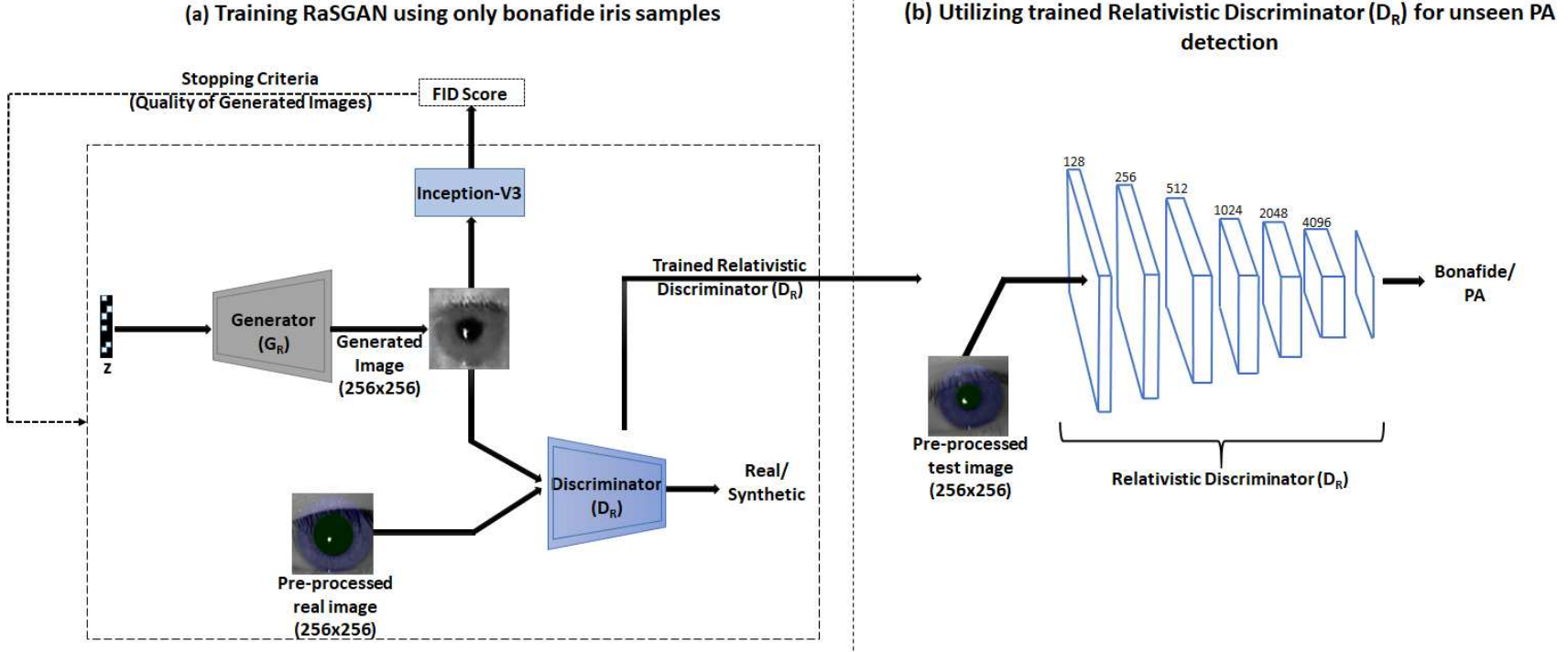}
\caption{Yadav \textit{et al.} \cite{yadav2020relativistic} proposed RaSGAN which was (a) trained to generate synthetic iris images. The discriminator network was trained to classify a given input iris sample as real or synthetic. (b) After training, the discriminator was exploited as a presentation attack detector to distinguish between bona fide and presentation attack iris samples. Thus, generative modelling to synthesize synthetic data benefits in improving model generalization on new and unseen test samples.}
\label{generative_modeling_synthetic}
\end{figure*}
    
    \subsection{Learning by synthesis}
    A machine learning model can be categorized as a discriminative or generative model. The former learns a conditional distribution $p(y|x;\theta)$, where $y$ denotes the output $y$ for the input sample $x$ and $\theta$ signifies model parameters. A generative model learns the joint distribution $p(x,y)$ and hence learns the distribution of data by learning to generate synthetic data. Such model is able to generalize on new and unseen test examples. A related seminal work \cite{wang2018hierarchical} presented %One such important study in human analysis includes the 
    a hierarchical generative model, % proposed by Wang \textit{et al.} . The authors propose a unified model that can 
    which jointly synthesizes eye images in a top-down approach, while estimating eye gaze in a bottom-to-up approach. A further generative modelling-based approach includes relativistic average standard generative adversarial network (RaSGAN) \cite{yadav2020relativistic} by Yadav \textit{et al.}. RaSGAN was trained to generate synthetic iris images, demonstrating the ability of its discriminator to generalize better on new and unseen presentation attacks (see Figure \ref{generative_modeling_synthetic}). Several approaches learning to synthesize data for improved model performance were proposed for re-identification of individuals \cite{an2017multi} \cite{chen2021joint} and face recognition \cite{gao2019learning} \cite{fu2021dvg}.
    
    % Gao \textit{et al.} \cite{gao2019learning} exploit synthetic face images to improve face recognition in uncontrolled scenarios with variation in poses. Given a face image, the model is trained to generate synthetic face with variation in pose. Afterwards, a contrastive loss is introduced in the Siamese network to perform training of the face recognition model. Fu \textit{et al.} \cite{fu2021dvg} propose DVG-Face framework to improve recognition of heterogeneous faces (g NIR-VIS, Sketch-Photo, Profile-Frontal Photo, Thermal-VIS, and ID-Camera). The authors propose a dual variational generator to generate synthetic heterogeneous face of the same identity. Later, the paired synthetic data is used to train the face recognition model using a contrastive loss.

\section{Open-Source Availability}
\label{sec:synthDatasets}

This Section provides an overview of synthetic datasets and synthetic data generation tools available for public usage. We emphasize the importance of sharing datasets and tools within the research community for improved reproducibility of results. That is, Table \ref{tab:synthData} presents publicly-available datasets comprised of synthetic data only. Further, Table \ref{tab:synthTools} introduces synthetic data generation tools to enable new researchers in the field of human analysis to build custom-generated datasets tailored to their needs.

 \begin{table*}
\label{table_example}
\begin{center}
\caption{Publicly available synthetic datasets. \label{tab:synthData}}
\resizebox{\linewidth}{!}{%
\begin{tabular}{|l|l|l|c|l|c|}
\hline
\textbf{Reference} & \textbf{Name} & \textbf{Application} & \textbf{Year} & \textbf{Data Type} & \textbf{Dataset Size}  \\ \hline\hline
%\cite{dou2021versatilegait} \textcolor{red}{?} & & Hand shape Recognition & & & 500,000  \\ \hline
%\cite{engelsma2022printsgan} \textcolor{red}{?} & & Fingerprint Recognition & & & 525,000 \\ \hline
%\cite{bouzaglo2022synthesis} \textcolor{red}{?} & & Fingerprint Recognition & &  & 100,000 pairs \\ \hline
%\cite{cao2018fingerprint} \textcolor{red}{?} & & Fingerprint Recognition & & & 40,000 \\ \hline

%\cite{mequanint2019weakly} \textcolor{red}{?} & & Eye closeness estimation & &  & 1.3 Million  \\ \hline

Wood \textit{et al.}~\cite{wood2021fake} & Microsoft Face Synthetics & Landmark localization, Face parsing & 2021 & Images & $100,000$ \\ \hline

Varol \textit{et al.}~\cite{varol2017learning} & SURREAL &  Human Pose Estimation & 2017 & Video Frames & $6,000,000$  \\ \hline

Barbosa \textit{et al.}~\cite{barbosa2018looking} & SOMASet & Person re-identification & 2017 & Images & $100,000$  \\ \hline

Varol \textit{et al.}~\cite{varol2021synthetic} & SURREACT &  Action Recognition & 2021 & Videos & $106,000$ \\ \hline

 Da \textit{et al.}~\cite{da2022dual} & Mixamo Kinetics &  Action Recognition & 2020 & Videos &  $36,195$ \\ \hline
 
 Ariz \textit{et al.}~\cite{ariz2016novel} & UPNA Synthetic Head Pose Database & Head Pose Estimation & 2016 & Videos &  $120$ \\ \hline
 
Roitberg\textit{et al.}~\cite{RoitbergSchneider2021Sims4ADL} & Sims4Action & Action Recognition & 2021 & Videos & $625.6$ minutes \\ \hline

Hwang\textit{et al.}~\cite{hwang2020eldersim} & KIST SynADL & Elderly Action Recognition & 2020 & Videos & $462,000$\\
\hline

\end{tabular}
}
\end{center}
\end{table*}

 \begin{table*}
\label{table_example}
\caption{Publicly available synthetic data generation models. \label{tab:synthTools}}
\begin{center}
\begin{tabular}{|l||l||c|l|}
\hline
\textbf{Reference} & \textbf{Application} & \textbf{Year} & \textbf{Method} \\ \hline \hline
%\cite{aranjuelo2021key} \textcolor{red}{?}  & people detection & 2021 & \\ \hline
%\cite{osadchy2017g}  \textcolor{red}{?}& Face generation tool &  & \\ \hline
%\cite{svoboda2020clustered} \textcolor{red}{?} & 3D hand shape recognition &  & \\ \hline
%\cite{han2018improving} \textcolor{red}{?}  & Face detection &  & \\ \hline
%\cite{dou2021versatilegait} \textcolor{red}{?}  & Gait Recognition &  & \\ \hline
%\cite{hillerstrom2014generation} \textcolor{red}{?} & Finger vein Generator tool &   & \\ \hline
%\cite{hillerstrom2014generation} \textcolor{red}{?} & gaze estimation, eye region generating tool &   &\\ \hline
%\cite{osadchy2017g} \textcolor{red}{?} & Face Authentication &  & \\ \hline
Drozdowski \textit{et al.}~\cite{drozdowski2017sic}  & Synthetic Iris Code Generator & 2017  & Handcrafted \\ \hline
%\cite{egger2018occlusion} \textcolor{red}{?} & 3D Morphable Model &  & \\ \hline
Li \textit{et al.}~\cite{li2017learning} & 3D Face Model Generation (FLAME) & 2019 & Handcrafted \\ \hline
Feng \textit{et al.}~\cite{feng2021learning} & 3D Face Model Registration (FLAME) & 2021 & Deep Neural Network \\ \hline
Colbois \textit{et al.}~\cite{colbois2021use} & Syn Multi-PIE Face Generation & 2021 & Deep Neural Network \\ \hline
Gerig \textit{et al.}~\cite{gerig2018morphable} & 3D Face Model Registration & 2018 & Handcrafted \\ \hline
Chan \textit{et al.}~\cite{chan2022efficient} & 3D Face Image Generation (EG3D) & 2022 & Deep Neural Network \\  \hline 
%\cite{guo2018face} \textcolor{red}{?} & Face Recognition with Eye Glasses &  & \\ \hline
Seneviratne \textit{et al.}~\cite{seneviratne2021multi} & Masked and unmasked Face Recognition & 2021 & Deep Neural Network \\ \hline
%\cite{wood2015rendering} \textcolor{red}{?} & Eye shape registration and gaze estimation & 2015 & \\ \hline
Karras \textit{et al.}~\cite{karras2020analyzing} & Face Image Generation (StyleGAN2) & 2020 & Deep Neural Network \\ \hline
Karras \textit{et al.}~\cite{karras2021alias} & Face Generation (StyleGAN3) & 2021 & Deep Neural Network \\ \hline
%\cite{hillerstrom2014generation} & Iris Image Generation & 2014 & Handcrafted \\ \hline
Maltoni \textit{et al.}~\cite{maltoni2009synthetic} & Fingerprint Image Generator (SFinGe) & 2009 & Handcrafted \\ \hline
 Sun \textit{et al.}~\cite{sun2019dissecting} & Person Re-Identification (PersonX) & 2019 & 3D Scenes and Models  \\ \hline
Hwang \textit{et al.}~\cite{hwang2020eldersim} & Elderly Action Recognition & 2020 & 3D Scenes and Models \\ \hline

\end{tabular}
\end{center}
\end{table*}

% \section{Include Figures of Synthetic samples}

\section{Challenges and Discussion}
We discussed benefits and means to generate and use synthetic datasets, placing emphasis on synthetic datasets being instrumental in mitigating challenges associated to real datasets. Despite related advances, there are a number of open research problems in this expanding field. %However, several open challenges are associated with synthetic datasets, which we will discuss next. 

\label{sec:challenges}
\begin{enumerate}

\item \textit{Identity leakage}. Studies that advocate using synthetic data for alleviating the privacy issues related to human data frequently do not conduct supporting experiments to show that there is no identity leakage from the training dataset \cite{tinsley2021face}. Such an assessment is critical to address privacy concerns related to sharing data for applications in human analysis. For instance, Engelsma \textit{et al.} \cite{engelsma2022printsgan} computed comparison scores between training samples and the synthetically generated fingerprints. Only $0.04\%$ of the training samples obtained comparison scores above a threshold, and all such samples were removed from the synthetic dataset before introducing it in the public domain. Similar practices need to be adopted by the research community working in human analysis to mitigate any identity leakage.

\item \textit{Lack of diversity}. The development of synthetic datasets in human analysis, generally speaking requires the generation of mated and non-mated samples. Recently, Grimmer \textit{et al.} \cite{grimmer2021generation} emphasised the challenge of approximating the full intra-identity variation of real datasets. Mated samples were obtained through minor manipulations of various semantic attributes in a given sample. However, the generated datasets still lacked diversity compared to real-world datasets. Another challenge has to do with creating synthetic datasets balanced \textit{w.r.t.} demographics. Generative models are often trained on biased datasets, thus lowering the generation quality of synthetic samples from underrepresented classes. We note that the current working draft of ISO/IEC 19795-10 \cite{ISO-IEC-WD-19795-10} aims at quantifying the biometric system performance variation across demographic groups, hence providing a standardized and consistent evaluation framework to assess the diversity of synthetic datasets.

\item \textit{Representation ability}. Numerous scientific work, particularly in biometrics \cite{zhang2021applicability, cao2018fingerprint}, have observed that while the generated synthetic data appears realistic, its characteristics represent notable differences from real biometric samples. Such observations question the representation ability of generated synthetic data and motivate the design of representative synthetic data generation methods.
For instance, synthetic videos (deepfakes) frequently incorporate artefacts \textit{e.g.}, in the eye or lip region. %, compared to real videos. Likewise, sometimes the 
In addition, characteristics/semantics in synthetic data differs from those in real samples. For instance, Gottschlich and Huckemann \cite{gottschlich2014separating} demonstrated the distribution of minutiae in synthetic fingerprints generated by SFinGe \cite{maltoni2009synthetic} was different from the one observed for real fingerprints. Therefore, the representation ability of synthetic data needs to be carefully validated before exploiting it for real-world applications.

 \begin{figure}
		\centering \includegraphics[width=0.45\textwidth]{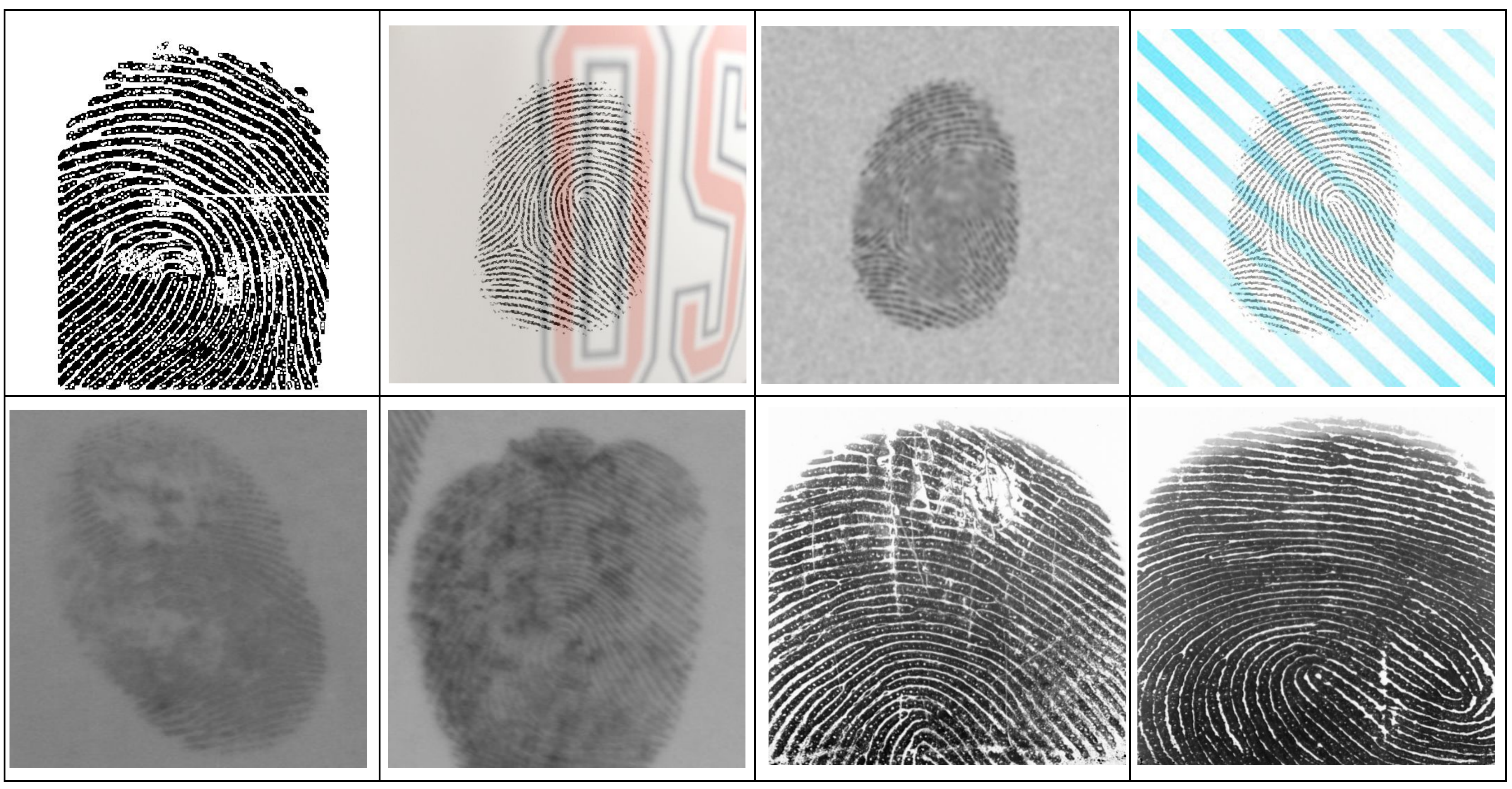}
		\caption{First row: synthetically distorted training samples. Second row: real testing samples for fingerprint enhancement algorithms, as used in \cite{joshi2021data}. The performance of the fingerprint enhancement model was directly dependent on how well the synthetic data modelled the noise observed in real fingerprints. Therefore, synthetically distorted training data must be publicly available to ensure fair comparison among different fingerprint enhancement algorithms.}
		\label{challenge_lack_comparison}
\end{figure}

\item \textit{Lack of comparison}. While scientific works in human analysis have been gradually exploiting methods for generating synthetic data, the related generated synthetic datasets are often not shared publicly. This is crucial, as the performance of human analysis models is directly dependent on how well synthetic data aligns with the testing dataset (see Figure \ref{challenge_lack_comparison}). In the case of Figure \ref{challenge_lack_comparison}, the fingerprint enhancement performance is dependent not only on the enhancement model but also on how carefully curated synthetic training data is. Therefore, to foster reproducibility and ensure a fair comparison among different methods, there is a need to share synthetic datasets publicly.

    % \item Leakage of training dataset into synthetic data:-- privacy issues
    % \item Synthetic data for public domain should address privacy concern
    % \item ARtifacts or representation ability -- how close to real data
    % \item diversity
    % \item Standard evaluation metrics 
    % \item VIDEO
    
\end{enumerate}

\section{Conclusions and Future Applications}
\label{sec:conclusion}
A review of the human analysis literature suggests that research in synthetic data is on the rise. This expansion is due to the large number of associated benefits in settings including
enrichment and replacement of existing real datasets. 
%These studies convincingly demonstrate the potential of synthetic data to mitigate issues such as privacy concerns, scalability, and generalization of unseen data. 

In this article, we reviewed some of the methods that have been developed for generation and exploitation of synthetic data in human analysis. In particular, we discussed techniques for generating semi-synthetic and fully synthetic data. 
Examples of related applications, we elaborated on include simulation of complex scenarios, mitigating bias and privacy concerns, increasing the size and diversity of training datasets, assessing scalability of systems, providing additional data for supervision, pre-training and fine-tuning of deep neural networks, enforcing consistency regularization, as well as adversarial attacks. Finally,
we discussed some of the open research problems in synthetic data research. 

%Furthermore, we discuss how synthetic data helps to enrich or replace an existing real dataset. These studies convincingly demonstrate the potential of 
We believe that synthetic data has the ability to mitigate issues related to privacy, scalability, and generalization of unseen data. Although so far synthetic data is abundantly utilized in human analysis, we believe that research directions including active learning, knowledge distillation and source-free domain adaptation will benefit in future from synthetic data. %are relatively unexplored for human analysis.

\ifCLASSOPTIONcompsoc
  % The Computer Society usually uses the plural form
  \section*{Acknowledgments}
\else
  % regular IEEE prefers the singular form
  \section*{Acknowledgment}
\fi

This research work has been supported by the French Government, by the National Research Agency (ANR) under Grant ANR-18-CE92-0024, project RESPECT, as well as by the
German Federal Ministry of Education and Research and the Hessian Ministry of Higher Education, Research, Science and the Arts within their joint support of the National Research Center for Applied Cybersecurity ATHENE.

% Can use something like this to put references on a page
% by themselves when using endfloat and the captionsoff option.
\ifCLASSOPTIONcaptionsoff
  \newpage
\fi

% trigger a \newpage just before the given reference
% number - used to balance the columns on the last page
% adjust value as needed - may need to be readjusted if
% the document is modified later
%\IEEEtriggeratref{8}
% The "triggered" command can be changed if desired:
%\IEEEtriggercmd{\enlargethispage{-5in}}

% references section

% can use a bibliography generated by BibTeX as a .bbl file
% BibTeX documentation can be easily obtained at:
% http://mirror.ctan.org/biblio/bibtex/contrib/doc/
% The IEEEtran BibTeX style support page is at:
% http://www.michaelshell.org/tex/ieeetran/bibtex/
%\bibliographystyle{IEEEtran}
% argument is your BibTeX string definitions and bibliography database(s)
%\bibliography{IEEEabrv,../bib/paper}
%
% <OR> manually copy in the resultant .bbl file
% set second argument of \begin to the number of references
% (used to reserve space for the reference number labels box)

% \begin{thebibliography}{1}

% \bibitem{IEEEhowto:kopka}
% H.~Kopka and P.~W. Daly, \emph{A Guide to \LaTeX}, 3rd~ed.\hskip 1em plus
%   0.5em minus 0.4em\relax Harlow, England: Addison-Wesley, 1999.

% \end{thebibliography}

\bibliographystyle{ieeetr}
\bibliography{ref}

\end{document}